\documentclass[lettersize,journal]{IEEEtran}
\usepackage{amsmath,amsfonts}
\usepackage{algorithmic}
\usepackage{algorithm}
\usepackage{array}
\usepackage{hyperref}
\usepackage{soul, color, xcolor}%高亮
\usepackage[caption=false,font=normalsize,labelfont=sf,textfont=sf]{subfig}
\usepackage{colortbl} %表格背景颜色
\usepackage{booktabs}       % 支持 \toprule \midrule \bottomrule
\usepackage{pifont}         % 支持 \ding{51} 和 \ding{55} 符号
\usepackage{textcomp}
\usepackage{stfloats}
\usepackage{url}
\usepackage{verbatim}
\usepackage{graphicx}
\usepackage{float}
\usepackage{cite}
\usepackage{orcidlink}
\usepackage{booktabs} % 用于美观的横线 (\toprule, \midrule, \bottomrule)
\usepackage{multirow} % 用于跨行单元格
\begin{document}

\title{Generative Multi-Focus Image Fusion}

\author{
Xinzhe Xie\,\orcidlink{0009-0008-6081-5654}\,$^{a}$,
Buyu Guo\,\orcidlink{0000-0001-6497-8540}\,$^{b,c,*}$,
Bolin Li\,\orcidlink{0009-0001-0650-3608}\,$^{a}$,
Shuangyan He\,\orcidlink{0000-0002-4147-1787}\,$^{a,c,d}$,
Yanzhen Gu\,\orcidlink{0000-0003-4616-867X}\,$^{a,c}$,
Qingyan Jiang\,$^{a}$,
Peiliang Li$^{a,c,d,*}$
\thanks{$^a$State Key Laboratory of Ocean Sensing \& Ocean College, Zhejiang University, Zhoushan, P. R. China, 316021. Emails: xiexinzhe@zju.edu.cn, libolin@zju.edu.cn, hesy@zju.edu.cn, guyanzhen@zju.edu.cn, jqy0610@zju.edu.cn, lipeiliang@zju.edu.cn.}%
\thanks{$^b$Donghai Laboratory, Zhoushan, P. R. China, 316021. Email: guobuyu@donghailab.com.}%
\thanks{$^c$Hainan Institute, Zhejiang University, Sanya, P. R. China, 572025.}%
\thanks{$^d$Hainan Provincial Observatory of Ecological Environment and Fishery Resource in Yazhou Bay, Sanya, P. R. China, 572025.}%
\thanks{*Corresponding authors: Buyu Guo (Tel: +86 17864270197, Email: guobuyu@donghailab.com) and Peiliang Li (Tel: +86 15964239596, Email: lipeiliang@zju.edu.cn).}%
\thanks{This work was supported by the Key R\&D Program of "Jianbing Lingyan+X" in Zhejiang Province (No. 2024SSYS0089), the Innovational Fund for Scientific and Technological Personnel of Hainan Province (No. KJRC2023D19), the Zhejiang Provincial Natural Science Foundation of China (No. LQN26D060019), and the Government in Guidance of Local Science and Technology Development (No. 2025ZY01111). Thanks are also due to the Hainan Observation and Research Station of Ecological Environment and Fishery Resource in Yazhou Bay for their support.}%
}

% \markboth{IEEE Transactions on Circuits and Systems for Video Technology,~Vol.~XX, No.~X, XXXX~2026}%
% {Xinzhe Xie \MakeLowercase{\textit{et al.}}: Generative Multi-Focus Image Fusion}

% \IEEEpubid{0000--0000/00\$00.00~\copyright~2026 IEEE}

\maketitle

\begin{abstract} 
Multi-focus image fusion aims to generate an all-in-focus image from a sequence of partially focused input images. Existing fusion algorithms generally assume that, for every spatial location in the scene, there is at least one input image in which that location is in focus. Furthermore, current fusion models often suffer from edge artifacts caused by uncertain focus estimation or hard-selection operations in complex real-world scenarios. To address these limitations, we propose a generative multi-focus image fusion framework, termed GMFF, which operates in two sequential stages. In the first stage, deterministic fusion is implemented using StackMFF V4, the latest version of the StackMFF series, and integrates the available focal plane information to produce an initial fused image. The second stage, generative restoration, is realized through IFControlNet, which leverages the generative capabilities of latent diffusion models to reconstruct content from missing focal planes, restore fine details, and eliminate edge artifacts. Each stage is independently developed and functions seamlessly in a cascaded manner. Extensive experiments demonstrate that GMFF achieves state-of-the-art fusion performance and exhibits significant potential for practical applications, particularly in scenarios involving complex multi-focal content.
The implementation is publicly available at \href{https://github.com/Xinzhe99/StackMFF-Series}{\textcolor{magenta}{https://github.com/Xinzhe99/StackMFF-Series}}.
\end{abstract}


\section{Introduction}
\IEEEPARstart{T}{he} aesthetic quality of portrait photography is often governed by bokeh, the characteristic soft blur that selectively appears in background regions while the primary subject remains sharply focused. Although bokeh enhances artistic imaging, out-of-focus regions in scientific and microscopic imaging lead to information loss, thereby impairing accurate scene analysis—a limitation that is particularly consequential in applications such as microscopy \cite{xie2024underwater}, biomedical diagnostics \cite{deng2025endoscopic}, and industrial chip inspection \cite{han2024defect}. Multi-focus image fusion (MFF) techniques address this challenge by integrating a series of images captured at distinct focal planes into a single, all-in-focus image, effectively recovering information that would otherwise be lost due to defocus.

Traditional MFF techniques can be broadly divided into two categories: spatial-domain-based \cite{liu2015general,nejati2015multi,rs15102486} and transform-domain-based \cite{li2024fractal,yang2007image,li2025multi} methods. These approaches rely on manually designed feature extraction and fusion rules, thereby exhibiting robust fusion performance. In recent years, advances in deep learning have shifted multi-focus image fusion from rule-driven to data-driven paradigms, significantly enhancing fusion performance. Moreover, leveraging the parallel computing capabilities of modern GPUs allows the processing time to be reduced to a real-time level \cite{xie2025lightmff}.

Despite the emergence of numerous state-of-the-art deep learning-based MFF techniques \cite{ZHENG2025102974,xie2024swinmff,zhang2021mff,cheng2023mufusion,xie2025multi,zhang2025ddbfusion} in recent years, these methods are generally restricted to the fusion of image pairs. The idea of achieving image stack fusion by iteratively applying image-pair fusion methods has seemingly become a consensus among researchers \cite{ma2021smfuse,guo2019fusegan}. We attribute this phenomenon to several factors: 1) this approach appears practically feasible; 2) benchmark datasets in this field primarily consist of image-pair examples; and 3) image-pair fusion is easier to implement than image-stack fusion. These factors collectively influence the prevailing research direction.

However, the StackMFF series \cite{xie2025stackmff,xie2025stackmffv2} has demonstrated through extensive experiments that existing methods deliver unsatisfactory fusion performance in processing large-scale multi-focus image stacks. This issue becomes particularly pronounced in stacks with more focal layers or for end-to-end fusion networks. The StackMFF series has proposed multiple solutions to this problem, which we discuss in detail in the Related Work section. In this paper, we further present the latest advancement of the series, StackMFF V4. It inherits the design principles of StackMFF V3, but through improvements in the network architecture, it achieves more effective intra-layer focus estimation and inter-layer information interaction, thereby improving image stack fusion performance with only one-fourth of the computational cost.

\begin{figure}[htbp]
    \centering
    \includegraphics[width=\linewidth]{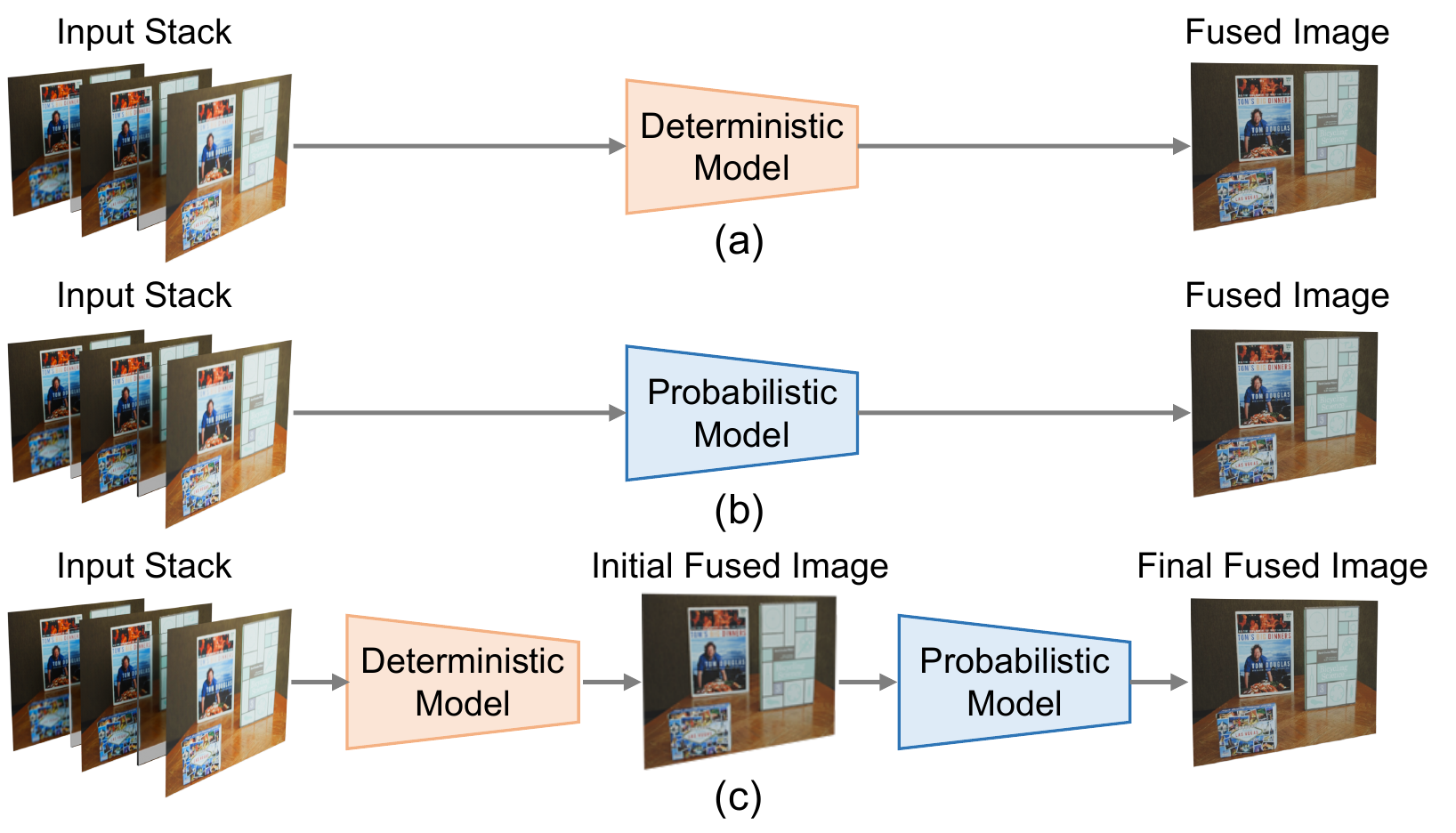}
    \caption{Comparison between prior models and our model (i.e., GMFF). 
    (a) Deterministic models applicable to image stack fusion, represented mainly by the StackMFF series; 
    (b) Representative methods that employ denoising probabilistic models for multi-focus image fusion, exemplified by FusionDiff \cite{li2024fusiondiff}, which require pairwise iterative fusion to achieve image stack fusion;
    (c) The proposed GMFF framework employs a deterministic model for pre-fusion, while the denoising probabilistic model is used for image restoration rather than for fusion, as shown in (b).}
    \label{fig:intro_1}
\end{figure}

In this work, we also investigate whether the quality of the fused image can, to some extent, be independent of the input stack’s quality and focal-plane completeness, as well as the performance of the employed fusion algorithm. Our motivation is grounded in two key observations. First, ideal multi-focus image stacks are rarely attainable; in practice, most captured stacks are suboptimal (criteria for an ideal stack are detailed in related work). As a result, the effective depth-of-field provided by real-world stacks may fail to cover the entire scene, potentially producing visibly blurred regions that degrade perceptual quality. Second, when fusing complex scenes, algorithmic limitations or operations resembling hard selection can introduce pronounced edge artifacts, yielding visually unappealing results. This challenge is particularly pronounced in decision-map-based fusion methods and tends to worsen as the number of focal planes increases. These observations naturally raise the question: if such issues are inherent, is it still possible to obtain reliable fused images with improved perceptual quality?

In recent years, denoising diffusion probabilistic models (DDPMs) \cite{ho2020denoising} have demonstrated remarkable performance in image generation. Methods such as DDRM \cite{DDRM2022}, DDNM \cite{wang2022zeroshotimagerestorationusing}, and GDP \cite{Fei_2023_CVPR} employ diffusion models as additional priors, conferring stronger generative capabilities than GAN-based approaches. Building upon these advances, DiffBIR \cite{DiffBIR} and UltraFusion \cite{ultrafusion} incorporate diffusion priors into image restoration and high-dynamic-range imaging, respectively, achieving state-of-the-art results. In this work, we investigate the integration of diffusion priors into multi-focus image fusion to improve fusion quality. Specifically, we decompose the fusion task into two stages, as illustrated in Fig. \ref{fig:intro_1}:

\begin{enumerate}
    \item \textit{Deterministic fusion}: In this stage, we employ multi-focus image fusion techniques to integrate the available focal plane information, producing an initial fused image. This preliminary fusion serves as the conditional input for the subsequent generative stage, ensuring fidelity and scene consistency. As existing methods have yet to achieve an optimal trade-off among generality, fusion quality, and computational efficiency, we introduce StackMFF V4, an efficient extension of StackMFF V3, which, to the best of our knowledge, represents the current state-of-the-art in overall performance for multi-focus image stack fusion.
    
    \item \textit{Generative restoration}: In the second stage, we leverage a pre-trained latent diffusion model to perform conditional image generation via ControlNet \cite{controlnet2023}, thereby restoring reliable details in regions that remain defocused after the initial fusion. This strategy also enhances fine structures and mitigates edge artifacts, which are challenging for conventional fusion methods to address, producing high-quality fused images that closely align with human visual perception.
\end{enumerate}

Notably, the second stage described herein is optional. Due to its generative nature, it may introduce uncertainty. Therefore, we recommend employing only the first-stage model in scientific imaging applications to mitigate potential reliability risks. The main contributions of this work can be summarized as follows:

\begin{itemize}
    \item We introduce the first generative multi-focus image fusion framework, GMFF, decoupling the MFF problem into two stages: deterministic fusion and generative restoration. This two-stage design enables GMFF to achieve state-of-the-art performance in multi-focus image fusion.
    
    \item We present the fourth-generation model of the StackMFF series, StackMFF V4. In this work, it serves as the fusion model in the deterministic fusion stage of GMFF, providing reliable conditional images for the generative restoration stage. StackMFF V4 is an improved version of StackMFF V3, incorporating three key enhancements: model scaling, a newly proposed Spatial Aggregation Cross-Layer Attention (SACA), and a newly introduced iterative refinement stage.
    
    \item We propose IFControlNet for the generative restoration stage, leveraging latent diffusion priors to reconstruct missing focal plane information with high fidelity. It also restores edge artifacts and enhances fine structures, thereby significantly improving the overall quality of the fused images.
\end{itemize}

In the following sections, we first review related works relevant to this study and then present the proposed GMFF framework in detail. Subsequently, extensive experiments are conducted to demonstrate the effectiveness of the GMFF framework. Finally, the paper concludes with a summary of our findings.

\section{Related work}
In this section, we first introduce the differences between our method and existing methods in terms of the requirements for the captured image stacks. Next, we review several existing MFF approaches, including image-pair fusion methods and image-stack fusion methods. Then, we present the generative models relevant to this work. Finally, we discuss recent advances in image restoration techniques that have been adapted for multi-focus image fusion.

\subsection{Requirements for the input image stack}
For almost all existing MFF algorithms, in order to obtain an all-in-focus image, it is necessary to ensure that every scene point is sharply focused in at least one image of the stack. The ideal acquisition strategy during capture should satisfy two criteria \cite{zhou2012focal}.
\begin{enumerate}
    \item Completeness: The collective depth of field (DoF) from all acquired images must encompass the entire target depth range. Given $DOF^*$ as the desired depth span, this condition can be expressed as:
    \begin{equation}
        DOF_1 \cup DOF_2 \cup DOF_3 \dots \cup DOF_n \supseteq DOF^*,
    \end{equation}
    where $\cup$ denotes a union operation and $\supseteq$ indicates a superset relationship.
    \item Efficiency: To minimize the total number of required images, individual depth-of-field (DoF) regions should not overlap. This can be represented as:
    \begin{equation}
        DOF_1 \cap DOF_2 \cap DOF_3 \dots \cap DOF_n = \emptyset,
    \end{equation}
    where $\cap$ denotes the intersection.
\end{enumerate}

In the method proposed in this paper, we no longer rely on the above criteria, particularly the completeness condition, and allow for some focal planes to be missing in the input image stack. This relaxation provides greater flexibility in image acquisition and makes our approach well-suited for \emph{quasi-static} scenes, where motion of dynamic objects may violate the commonly assumed static scene requirement for image stack fusion, such as in biological microscopy \cite{xie2024underwater}. Consequently, the number of exposures and refocusing operations is reduced, better aligning with the assumption of static image stacks and enhancing practical applicability in scenarios where strict static conditions cannot be guaranteed.

\subsection{Multi-focus image fusion}
Traditional MFF algorithms can generally be divided into spatial-domain methods \cite{liu2015general,nejati2015multi,rs15102486} and transform-domain methods \cite{li2024fractal,yang2007image,li2025multi}. Spatial-domain methods compute decision maps based on local activity measures and fuse images accordingly, whereas transform-domain methods first map the images into a transform domain, integrate information across the focal planes, and then reconstruct the all-in-focus image. Moreover, deep learning has recently opened new avenues for multi-focus fusion, primarily as end-to-end approaches that directly map input images to an all-in-focus image \cite{lin2024fusion2void,xie2024swinmff,cheng2023mufusion} and decision-map-based approaches that estimate pixel- or region-level focus weights prior to fusion \cite{xie2025multi,quan2025multi,hu2023zmff}.

Recently, researchers have begun to extend learning-based image-pair fusion paradigms to image-stack fusion, aiming to enhance the applicability of fusion algorithms in real-world scenarios \cite{xie2025stackmff,xie2025stackmffv2,araujo2023towards,li2023generation}. Araujo et al. \cite{araujo2022focus} use pseudo ground truth generated by commercial focus stacking software 
\href{https://www.heliconsoft.com/heliconsoft-products/helicon-focus/}{\textcolor{magenta}{Helicon Focus}} as supervisory images, training networks on a small captured dataset. Since these supervisory images are not true ground truth and the training data are limited, the fusion performance is constrained. GRFusion \cite{li2023generation} detects the focus attributes of each source image and introduces a hard-pixel-guided recombination mechanism to fuse arbitrary input images. Although it claims to support fusion of arbitrary numbers of inputs, its implementation in practice is limited to a few predefined input quantities. Consequently, GRFusion functions more as a task-specific multi-image fusion system tailored for particular experimental settings, rather than a truly general-purpose framework capable of handling arbitrary input counts.

The StackMFF series \cite{xie2025stackmff,xie2025stackmffv2} represents the most comprehensive work in this field to date. Its first version, StackMFF \cite{xie2025stackmff}, is an end-to-end fusion network based on 3D convolutional neural networks, which demonstrated the potential of learning-based stack fusion networks through validation on a large-scale synthetic dataset. The second version, StackMFF V2 \cite{xie2025stackmffv2}, reformulates image-stack fusion as a focal-plane depth regression problem. It leverages the numerical equivalence between depth maps and focus maps, applicable when synthetic image stacks for training are generated by linear layering according to depth maps. This version proposes a novel method that allows training MFF networks using depth maps. The third version, StackMFF V3, formulates the image-stack fusion task as a pixel-level classification problem, representing the current state-of-the-art solution in terms of fusion performance, functionality, and overall generality. In this work, we present the fourth version of the StackMFF series, StackMFF V4. Building upon the fusion paradigm established in StackMFF V3, StackMFF V4 enhances fusion quality while reducing inference time to approximately one-quarter of that of its predecessor.

\subsection{Generative models}
In recent years, diffusion models have rapidly emerged \cite{ho2020denoising,nichol2021improved,song2020denoising}, finding successful applications in a wide range of vision tasks, including controllable image generation, image restoration, image editing, and image synthesis. However, in the context of image fusion, their application has largely been limited to conceptual frameworks. For instance, FusionDiff \cite{li2024fusiondiff} operates without leveraging diffusion priors learned from large-scale datasets, which limits generalization and often produces fused images with noticeable color deviations. Mask-DiFuser \cite{tang2025mask} reformulates image fusion as an unsupervised dual-masked reconstruction task using a fully self-trained masked diffusion process. Although DDFM \cite{zhao2023ddfm} incorporates diffusion priors, these methods remain limited to two-image fusion. In contrast, our framework supports multi-image stacks as input and leverages pre-trained diffusion priors to generate reliable details in regions that remain defocused after the initial deterministic fusion, simultaneously removing edge artifacts and enhancing fine structures.

\subsection{Image restoration}
Blind image deblurring \cite{quan2021gaussian,lee2021iterative,ruan2022learning,zamir2022restormer,ruan2023revisiting,quan2023single,quan2023neumann,zhai2024efficient} aims to recover a sharp image from a single blurred observation without access to explicit imaging information, thereby effectively inverting the degradation process. This task is fundamentally different from MFF, which fuses multiple images captured at different focal planes to produce an all-in-focus image. One might intuitively attempt to improve the perceptual quality of fused images by applying existing blind deblurring techniques to remove residual blur. However, our extensive experiments demonstrate the ineffectiveness of this approach. The underlying reason is that, in our setting, most regions of the input images are already sharp, whereas conventional deblurring methods generally assume that the entire image is degraded by blur.

Recently, techniques from image restoration and inpainting have been adapted to MFF to enhance fine details and reduce artifacts. Specifically, Wang et al.~\cite{WANG2024104473} formulate MFF as a two-stage process. First, an initial decision map is generated using a traditional local activity measure. Subsequently, an image restoration network refines this map under the guidance of input image gradients, enhancing edge details and correcting prominent fusion errors. In contrast, Fusion2Void~\cite{lin2024fusion2void} treats MFF as an inpainting task, regarding out-of-focus regions as \emph{voids} to be reconstructed from the surrounding in-focus context. This formulation enables unsupervised fusion by leveraging inpainting-based supervision.

Unlike these approaches, our GMFF framework adopts a two-stage paradigm comprising deterministic fusion followed by generative restoration. Rather than relying on restoration or inpainting to supervise image fusion, GMFF leverages a pre-trained diffusion model as a robust generative prior. This framework directly operates on the initially fused image, performing generative restoration on regions that remain defocused and on edge artifacts, thereby effectively addressing both residual blur and artifact issues.

\section{Proposed Method}
In this section, we first briefly introduce the proposed GMFF framework. It consists of two stages: 1) the deterministic fusion stage and 2) the generative restoration stage. Subsequently, we provide a detailed description of each stage.

\subsection{Method overview}
\begin{figure*}[htbp] % 
    \centering
    \includegraphics[width=0.9\linewidth]{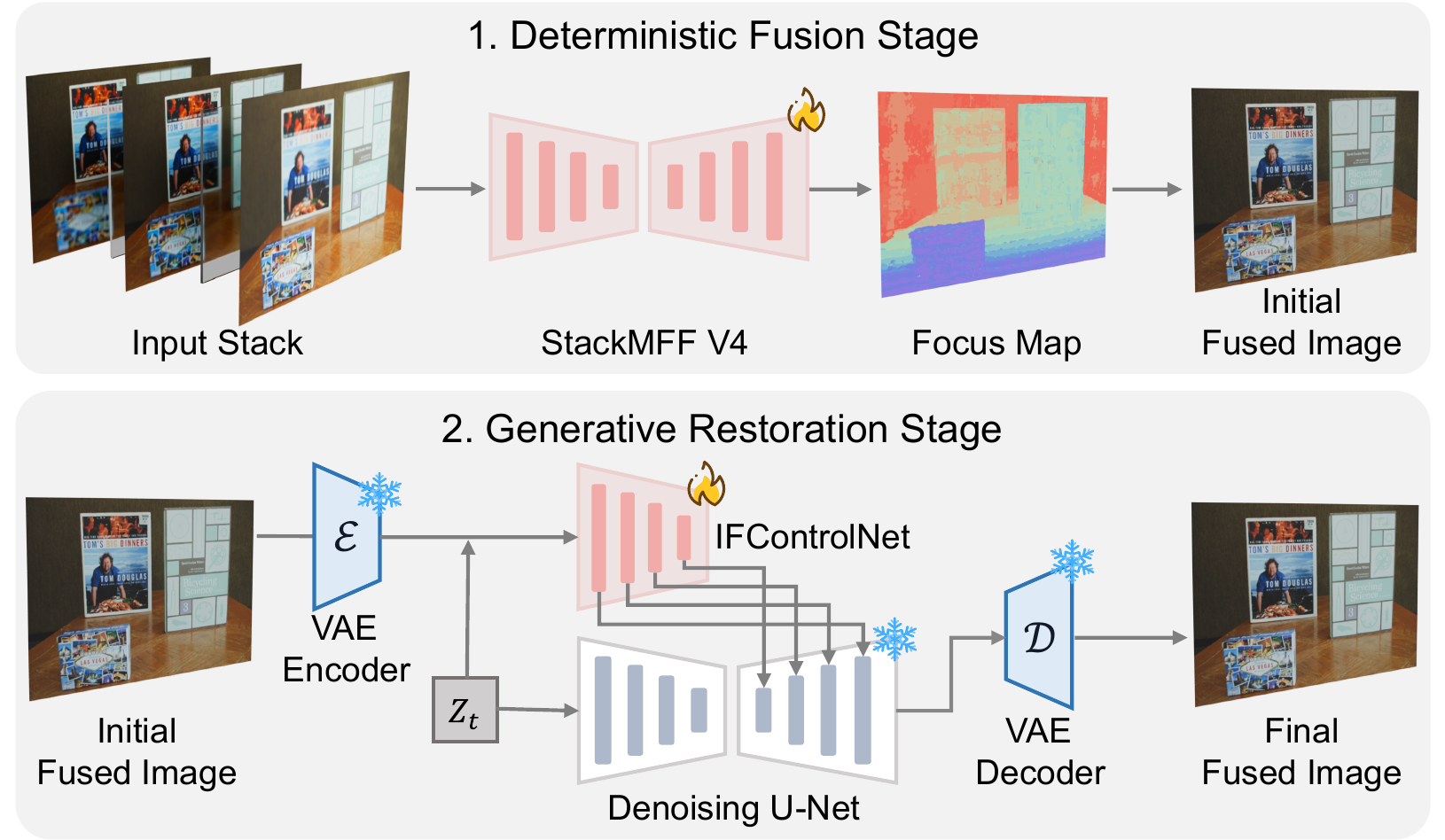}
    \caption{Overview of the proposed generative multi-focus image fusion framework (GMFF), which consists of two stages: deterministic fusion and generative restoration.}
    \label{fig:framework}
\end{figure*}

Fig. \ref{fig:framework} illustrates the proposed two-stage multi-focus image fusion framework (GMFF). In the first stage, we employ the fourth-generation model of the StackMFF series, StackMFF V4, for the deterministic fusion stage of GMFF. It integrates all available focal plane information from the input image stack efficiently and produces a reliable initial fused image.

In this work, our goal is not only to construct a more effective image stack fusion network but also to leverage a powerful generative prior to address the issues of missing focal planes and the fusion artifacts. By incorporating conditional inputs, such as edge maps and depth maps, generative diffusion priors demonstrate their effectiveness in conditional image generation \cite{controlnet2023}. This approach thus provides a potential solution to the aforementioned issues. 

Building on the first stage, we introduce a second stage for image restoration, in which the initial fused image serves as a conditional input to the diffusion model. The diffusion prior is then employed to refine the initial fused image and generate the final image. These two stages are decoupled and optimized independently. Thus, any existing fusion model can be seamlessly integrated into our framework. This two-stage pipeline provides a flexible, stable, and unified solution to the multi-focus image fusion problem.

\subsection{Deterministic fusion}
\begin{figure*}
    \centering
    \includegraphics[width=\linewidth]{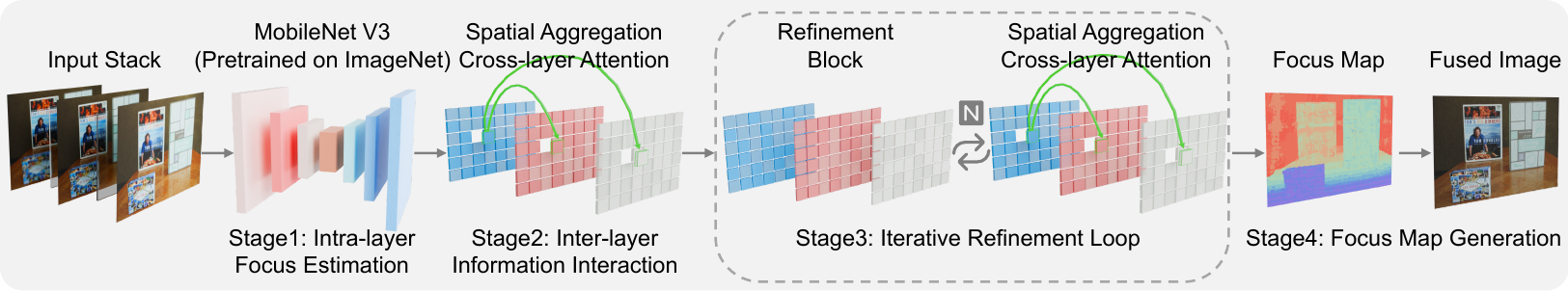}
    \caption{Framework of the proposed StackMFF~V4.}
    \label{fig:stackmffv4_framework}
\end{figure*}

In the deterministic fusion stage, the goal is to accurately integrate the available focal-plane information into the initial fused image. This is crucial because the diffusion model in the second stage is highly sensitive to the conditional input, which means that without precise fusion, the resulting fused image may deviate significantly from the true scene. Therefore, the choice of the deterministic model for this stage is critical. Although numerous deep learning-based image-pair fusion methods have been proposed, they are insufficient for the image stack fusion task. Errors tend to accumulate, thereby hindering the generation of reliable fused images. In contrast, traditional algorithms are generally robust for image stack fusion tasks and can be considered viable alternatives. Nevertheless, StackMFF V4 is adopted as the deterministic model for this stage. As the fourth iteration of the StackMFF series, StackMFF V4 inherits the fusion paradigm of StackMFF V3 and introduces architectural enhancements. It achieves state-of-the-art fusion performance while maintaining the lowest computational cost and the fastest inference among image stack fusion methods.

Fig. \ref{fig:stackmffv4_framework} illustrates the framework of StackMFF V4, which extends the StackMFF V3 fusion framework by incorporating an additional stage—the iterative refinement loop—before the focus map generation stage. As a result, StackMFF V4 consists of four stages in total: 1) \textit{intra-layer focus estimation}; 2) \textit{inter-layer information interaction}; 3) \textit{iterative refinement loop}; and 4) \textit{focus map generation}. Due to space limitations, only the key improvements in this version are described, and readers are referred to the original StackMFF V3 paper for further details.

\subsubsection{Scaling up the model}
The first improvement concerns the intra-layer focus estimation stage. StackMFF V2 \cite{xie2025stackmffv2} employs a specifically designed lightweight ULDA-Net to independently estimate the focus level within each layer. To balance the trade-off between receptive field size and computational cost, StackMFF V3 replaces the ULDA-Net with PFMLP \cite{huangpfmlp}, an MLP-based architecture that serves as the intra-layer focus estimation network. Owing to the more computationally efficient inter-layer modeling approach introduced in the following section, a larger computational budget can be allocated to the intra-layer focus estimation stage, thereby enhancing the model’s ability to discriminate focused regions. In StackMFF V4, MobileNetV3-Large \cite{howard2019searching} is employed as the backbone of the intra-layer focus estimation stage. The encoder is initialized with weights pretrained on ImageNet \cite{imagenet2009}, which accelerates convergence and enhances feature extraction. This improvement enables more accurate focus estimation, even under limited training data.

\subsubsection{Spatial Aggregation Cross-Layer Attention}
\begin{figure}[htbp]
    \centering
    \includegraphics[width=\linewidth]{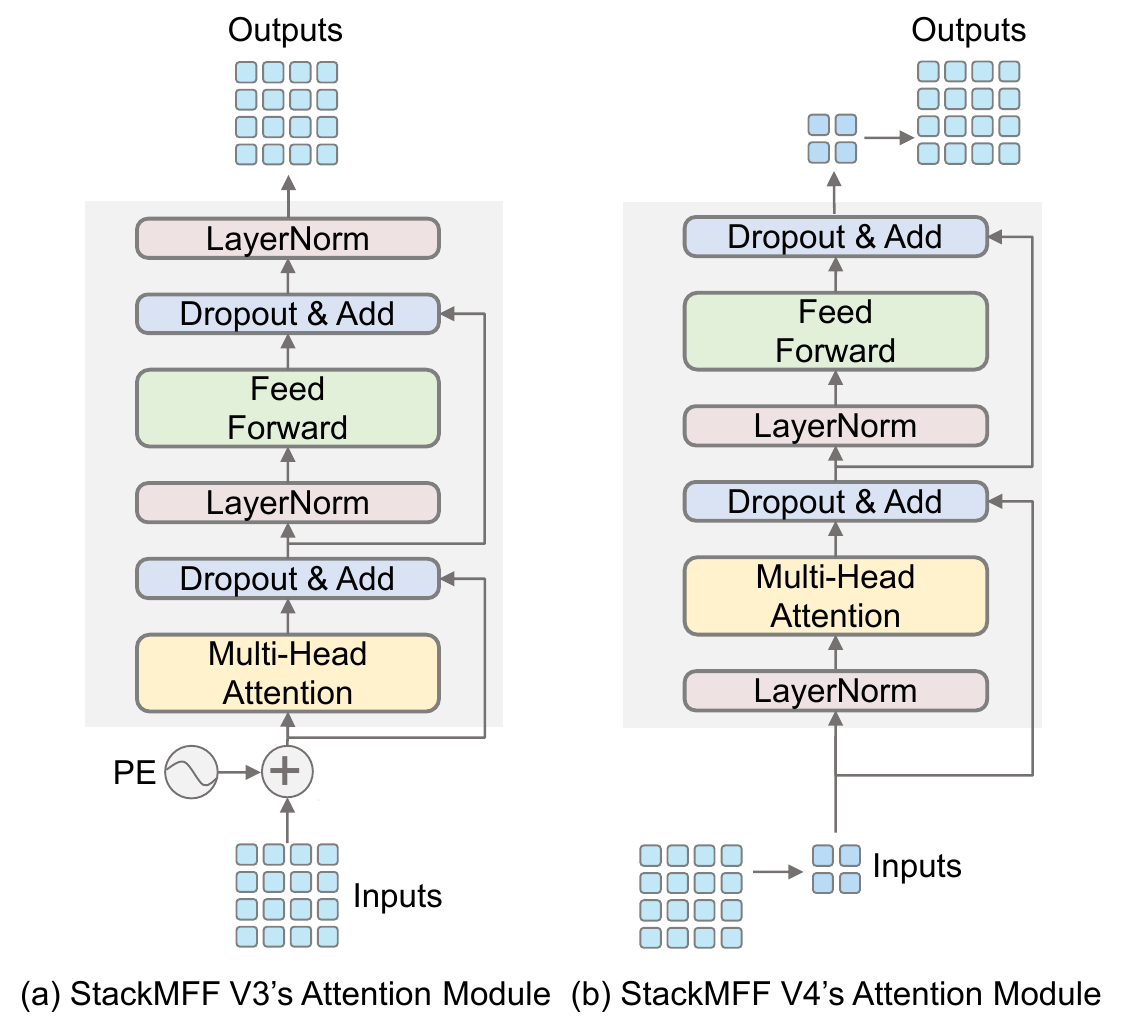}
    \caption{Detailed transformer module comparison between StackMFF V3 (Pixel-wise Cross-Layer Attention, PCA) and StackMFF V4 (Spatial Aggregation Cross-Layer Attention, SACA).}
    \label{fig:stackmffv4_attention}
\end{figure}

In StackMFF V3, inter-layer information interaction is modeled via a self-attention mechanism along the depth dimension, referred to as Pixel-wise Cross-Layer Attention (PCA). Since the number of layers is much smaller than the image resolution, the quadratic complexity of self-attention, \(O(N^2)\), remains computationally tractable at the per-pixel level. However, the empirical success of Efficient LoFTR~\cite{wang2024efficient} suggests that aggregating features along the spatial dimension before performing self-attention constitutes a more effective strategy. Such aggregation enables inter-layer interactions to encode contextual pixel-level information, thereby enhancing modeling efficacy while substantially reducing computational overhead.

As illustrated in Fig.~\ref{fig:stackmffv4_attention}, in the inter-layer information interaction stage of StackMFF V4, average pooling is first applied along the spatial dimensions before feeding the features into a Transformer block. This approach is referred to as Spatial Aggregation Cross-Layer Attention (SACA). Bilinear interpolation is subsequently employed to restore the original resolution. Additionally, the placement of normalization layers is adjusted to better suit visual processing tasks, and the rotary positional encoding is removed, as the network is designed to be invariant to layer ordering. The dual benefits of these modifications are analyzed in detail in the ablation study.

\subsubsection{Iterative refinement}
In StackMFF V4, a novel iterative refinement loop stage is introduced. This design is motivated by the insight that StackMFF V3 operates as a causal network. The quality of inter-layer modeling is contingent upon the accuracy of intra-layer focus estimation, which subsequently influences both the focus maps and the fused image. After employing Spatial Aggregation Cross-Layer Attention (SACA), it is observed that the computational cost of inter-layer modeling is significantly reduced, while modeling performance is further enhanced. This efficiency enables the introduction of an iterative refinement loop before the focus map generation stage, while maintaining a controlled computational budget. Each loop consists of a refinement block for further intra-layer focus estimation and a SACA module to facilitate inter-layer information exchange. Each loop leverages the results of inter-layer interactions to iteratively refine intra-layer focus estimation, thereby enhancing both intra-layer and inter-layer modeling efficacy. Experiments show that even a single iteration achieves substantial performance gains in StackMFF V4.

Experimental results show that, with these three improvements, StackMFF V4 achieves state-of-the-art performance in multi-focus image stack fusion tasks, while maintaining the lowest computational cost and fastest inference. These results establish StackMFF V4 as the most advanced model in the field, providing reliable conditional images for the subsequent generative restoration stage.

\subsection{Generative restoration}
Several recent studies \cite{DiffBIR,lin2025harnessing,ultrafusion} have demonstrated that diffusion model priors, after large-scale pretraining, hold great potential for addressing residual blur, edge artifacts, and detail loss. In this section, we describe the Stable Diffusion 2 \cite{rombach2022high} and illustrate how ControlNet \cite{controlnet2023} achieves scene-consistent and reliable generative restoration.

\subsubsection{Preliminary}
The proposed method builds upon the large-scale latent diffusion model, \href{https://github.com/Stability-AI/stablediffusion}{\textcolor{magenta}{Stable Diffusion 2.1-base}}. To enhance training stability and computational efficiency, Stable Diffusion initially pretrains a variational autoencoder (VAE)~\cite{kingma2013auto} to map an image $x$ to a latent code $z = \mathcal{E}(x)$ via an encoder $\mathcal{E}$ and to reconstruct it through a decoder $\mathcal{D}$. Both diffusion and denoising are performed in this latent space.

During the diffusion process, Gaussian noise with variance $\beta_t \in (0,1)$ is incrementally added to the latent code at each time step $t$:
\begin{equation}
z_t = \sqrt{\bar{\alpha}_t} z + \sqrt{1 - \bar{\alpha}_t} \epsilon, \quad \epsilon \sim \mathcal{N}(0, I),
\label{eq:diffusion_process}
\end{equation}
where $\alpha_t = 1 - \beta_t$ and $\bar{\alpha}_t = \prod_{s=1}^{t} \alpha_s$. As $t$ increases, $z_t$ approaches a standard Gaussian distribution. A denoising network $\epsilon_\theta$ is trained to predict $\epsilon$ conditioned on a text prompt $c$ and a randomly sampled timestep $t$ by optimizing the following objective function:
\begin{equation}
\mathcal{L}_{\text{ldm}} =
\mathbb{E}_{z,c,t,\epsilon}
\Big\|
\epsilon - \epsilon_\theta
\big(
\sqrt{\bar{\alpha}_t} z + \sqrt{1 - \bar{\alpha}_t} \epsilon, c, t
\big)
\Big\|_2^2.
\label{eq:ldm_loss}
\end{equation}

\subsubsection{IFControlNet}
Using the deterministic fusion output $I_{IF}$ as the conditional input, we introduce IFControlNet for generative restoration. The overall process consists of four key components: (1) latent representation, (2) conditional network, (3) latent denoising, and (4) image reconstruction.

\textbf{1) latent representation.}  
A VAE encoder $\mathcal{E}$ maps $I_{IF}$ into the latent space, yielding $c_{IF} = \mathcal{E}(I_{IF})$. This latent encoding captures a compact yet informative representation of the fused image, making it suitable for guiding the generative process.

\textbf{2) conditional network.}  
To inject $c_{IF}$ into the generative process, we propose IFControlNet, a ControlNet variant specifically designed for the GMFF framework. This network is primarily used to remove residual blur and fusion artifacts from the deterministic fused image. Given the noisy latent $z_t$ and the conditional latent $c_{IF}$, IFControlNet produces a residual $\Delta z_t$ that is added to the input of the denoising U-Net:
\begin{equation}
\tilde{z}_t = z_t + f_\phi(z_t, c_{IF}, t).
\end{equation}
By incorporating this conditional residual, IFControlNet ensures that the structural details in $I_{IF}$ are effectively preserved during denoising while mitigating residual blur and artifacts.

\textbf{3) latent denoising.}  
The conditionally modulated latent $\tilde{z}_t$ is iteratively denoised by the U-Net:
\begin{equation}
z_{t-1} = \frac{1}{\sqrt{\alpha_t}} \Big( \tilde{z}_t - \frac{1-\alpha_t}{\sqrt{1-\bar{\alpha}_t}} \epsilon_\theta(\tilde{z}_t, t) \Big) + \sigma_t \epsilon',
\end{equation}
where $\epsilon'$ is Gaussian noise. By incorporating the conditional residual, the latent is guided to align with $I_{IF}$ while being enriched with realistic textures.

\textbf{4) image reconstruction.}  
Finally, the refined latent $z_0$ is decoded via $\mathcal{D}$:
\begin{equation}
\hat{I} = \mathcal{D}(z_0),
\end{equation}
producing an output image that preserves the structural fidelity of the fused image while introducing generative enhancements.

\section{Experiments}
In this section, we first describe the implementation details of the GMFF framework and the associated evaluation protocol. Subsequently, we conduct extensive quantitative and qualitative evaluations, along with efficiency analyses, for both StackMFF V4 and the complete GMFF framework. These experiments demonstrate the feasibility and effectiveness of our approach. In addition, we perform a comprehensive set of ablation studies to highlight the advantages of StackMFF V4 over its predecessors.

\subsection{Implementation details}

\subsubsection{Training Strategy} 
All experiments were conducted on a high-performance computing platform equipped with dual NVIDIA A6000 GPUs and an Intel(R) Xeon(R) Platinum 8375C CPU. Training was performed using both GPUs, while inference utilized only a single GPU.

For training the GMFF network, images were sourced from DUTS \cite{wang2017learning}, NYU Depth V2 \cite{silberman2012indoor}, DIODE \cite{vasiljevic2019diode}, Cityscapes \cite{cordts2016cityscapes}, and ADE20K \cite{zhou2019semantic}. The monocular depth estimation network Depth Anything V2 \cite{yang2024depth} was used to obtain the depth maps required for synthesizing multi-focus image stacks. For all datasets except NYU Depth V2, which already provides high-quality depth maps, scene depths were estimated using Depth Anything V2. Subsequently, realistic multi-focus image stacks were rendered via a depth-based linear stratification method, producing corresponding focus maps for training StackMFF V4. The trained StackMFF V4 was then used to process all synthesized multi-focus stacks, generating input images for training IFControlNet. To simulate missing focal planes and local detail loss, $0\%-50\%$ of the input layers were randomly discarded during training.

StackMFF V4 was trained using the AdamW optimizer with a batch size of 12 and an initial learning rate of $1 \times 10^{-3}$, which was exponentially decayed by a factor of 0.9 per epoch. Both the synthesized multi-focus stacks and the fully sharp supervision images were resized to a resolution of $384 \times 384$. The training datasets were the same as those used for StackMFF V3. However, leveraging pretraining on ImageNet, only one-tenth of the StackMFF V3 training data was randomly sampled to achieve convergence. Training lasted 50 epochs and took approximately 8 hours, with early stopping applied.

For IFControlNet training, the weights of the VAE and Stable Diffusion 2.1-base were frozen, and only the conditional network was updated. To improve training efficiency, the network was initialized with the IRControlNet weights from DiffBIR \cite{DiffBIR} and then fully fine-tuned. AdamW was used as the optimizer with a batch size of 8. During the first 30,000 steps, the learning rate was set to $3 \times 10^{-6}$ based on the linear scaling principle from DiffBIR; for the subsequent 20,000 steps, the learning rate was reduced to $3 \times 10^{-7}$. Training IFControlNet took approximately 16 hours. Inputs to IFControlNet were generated using the trained StackMFF V4, while the corresponding ground truth was used for supervision.

\subsubsection{Datasets for evaluation}
The benchmark datasets used in StackMFF~V2 \cite{xie2025stackmffv2}, including Mobile Depth~\cite{suwajanakorn2015depth}, FlyingThings3D~\cite{mayer2016large}, Middlebury Stereo~\cite{scharstein2014high}, and Road-MF~\cite{li2024samf}, were employed to evaluate the proposed GMFF fusion framework. These datasets cover a wide range of indoor and outdoor scenes, including both real-world and synthetic images. All four datasets were employed in the deterministic fusion stage of StackMFF~V4. For the evaluation of the generative restoration stage, however, the synthetic FlyingThings3D dataset and the Road-MF dataset—both exhibiting domain-specific bias (i.e., predominantly containing a single type of image)—were excluded from evaluation.

\subsubsection{Methods for comparison and evaluation metrics}
Given that the two stages address different tasks, we evaluate the fusion results of each stage separately to provide a more comprehensive analysis. 

For the deterministic fusion stage, we compare it with 18 representative state-of-the-art MFF methods. These include five traditional techniques—CVT~\cite{guo2012multifocus}, DWT~\cite{li1995multisensor}, DCT~\cite{HAGHIGHAT2011789}, DTCWT~\cite{lewis2007pixel}, and NSCT~\cite{yang2007image}—and 13 learning-based approaches: IFCNN~\cite{zhang2020ifcnn}, U2Fusion~\cite{xu2020u2fusion}, SDNet~\cite{zhang2021sdnet}, MFF-GAN~\cite{zhang2021mff}, SwinFusion~\cite{ma2022swinfusion}, MUFusion~\cite{cheng2023mufusion}, SwinMFF~\cite{xie2024swinmff}, DDBFusion~\cite{zhang2025ddbfusion}, CCSR-Net / MCCSR-Net~\cite{ZHENG2025102974}. Additionally, three StackMFF series models specifically designed for image stack fusion are included: StackMFF~\cite{xie2025stackmff}, StackMFF~V2~\cite{xie2025stackmffv2}, and StackMFF~V3. For comparison, two commercial focus-stacking software packages that support batch processing are also considered: Helicon Focus~8 (offering three fusion methods) and Zerene Stacker (offering two).

We adopt reference-based evaluation metrics, including SSIM and PSNR, to quantitatively assess the fusion quality of the resulting all-in-focus images. SSIM (Structural Similarity Index) measures the perceptual similarity between the fused image and the ground truth, considering luminance, contrast, and structural similarity. Higher SSIM values thus indicate better preservation of structural information. PSNR evaluates the pixel-wise reconstruction fidelity by computing the logarithmic ratio between the maximum possible signal power and the mean squared error (MSE). Higher PSNR values correspond to greater reconstruction accuracy of the fused image. These metrics are widely used in multi-focus image fusion research, enabling standardized and objective comparisons across different fusion methods.

For the generative restoration stage, since there is currently no existing model similar to GMFF in the MFF field that performs generative restoration and optimization on the fusion results, we compare our approach with eight state-of-the-art blind image restoration networks specifically designed for defocus deblurring, listed in chronological order: GKMNet \cite{quan2021gaussian}, IFAN \cite{lee2021iterative}, DRBNet \cite{ruan2022learning}, Restormer \cite{zamir2022restormer}, LaKDNet \cite{ruan2023revisiting}, INIKNet \cite{quan2023single}, NRKNet \cite{quan2023neumann}, and DEDDNet \cite{zhai2024efficient}.

Although ground-truth images are available, the generative restoration stage of GMFF enhances perceptual quality and optimizes local details, which may not necessarily produce outputs that are closer to the ground truth in a pixel-wise sense. Consequently, reference-based metrics, such as SSIM and PSNR, tend to underestimate the quality of the generated images. Therefore, we adopt the no-reference metrics BRISQUE \cite{mittal2012no} and PIQE \cite{venkatanath2015blind} to quantitatively evaluate the perceptual quality of the restored results. BRISQUE is a no-reference metric based on natural scene statistics (NSS), which quantifies the deviation of an image’s statistical properties from those of typical high-quality natural images. PIQE is another no-reference metric that estimates image quality by computing distortion measures within local spatial regions. It evaluates the level of perceptually significant distortion, assigning lower scores to images with fewer visible artifacts, thereby indicating better perceptual quality. Together, BRISQUE and PIQE provide a comprehensive assessment of the perceptual and structural fidelity of the generated restoration results in the absence of ground-truth references, with lower scores corresponding to better perceptual quality.

\subsection{Evaluation of StackMFF V4}
\subsubsection{Qualitative comparison}
\begin{figure*} 
    \centering
    \includegraphics[width=\linewidth]{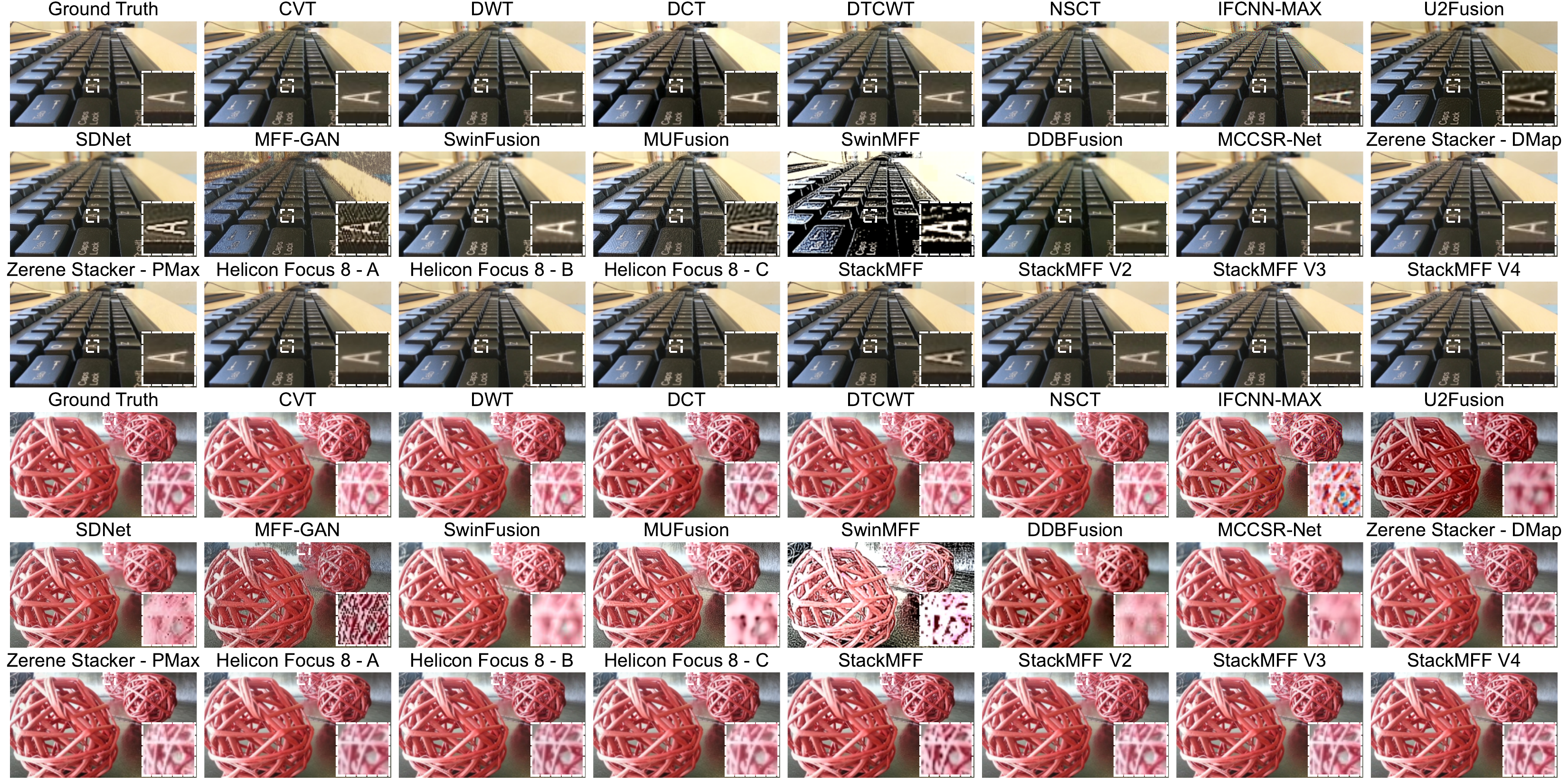}
    \caption{Comparison of fusion results produced by various methods on the Mobile Depth dataset \cite{suwajanakorn2015depth}. The examples correspond to ``keyboard'' and ``balls'', respectively.}
    \label{fig:stackmffv4_compare_methods_mobile_depth}
\end{figure*}

Fig. \ref{fig:stackmffv4_compare_methods_mobile_depth} presents the fusion results of various MFF algorithms on the Mobile Depth dataset. The magnified regions in the bottom-right corners of each subfigure highlight specific local details. It can be observed that all traditional methods, including CVT, DWT, and DCT, as well as several commercial focus-stacking software packages, produce high-quality fusion results closely approximating the ground truth. In contrast, most deep learning-based multi-focus image fusion networks struggle to generalize effectively to image stack fusion tasks. For instance, in the first ``keyboard'' example, U2Fusion, SDNet, MFF-GAN, SwinMFF, and MUFusion exhibit pronounced noise and artifacts. In the second ``balls'' example, in addition to noise, U2Fusion, SwinFusion, DDBfusion, and MCCSR-Net exhibit evident fusion failures, with the magnified regions appearing blurred. In comparison, the StackMFF series, although also learning-based, demonstrates a clear advantage over these pairwise fusion networks for image stack fusion tasks. Moreover, the magnified regions indicate that the fusion quality progressively improves from StackMFF to StackMFF V4, yielding highly satisfactory results. Furthermore, StackMFF V4, as the most computationally efficient model in the StackMFF series, achieves the highest fusion quality within the series.

\begin{figure*} 
    \centering
    \includegraphics[width=\linewidth]{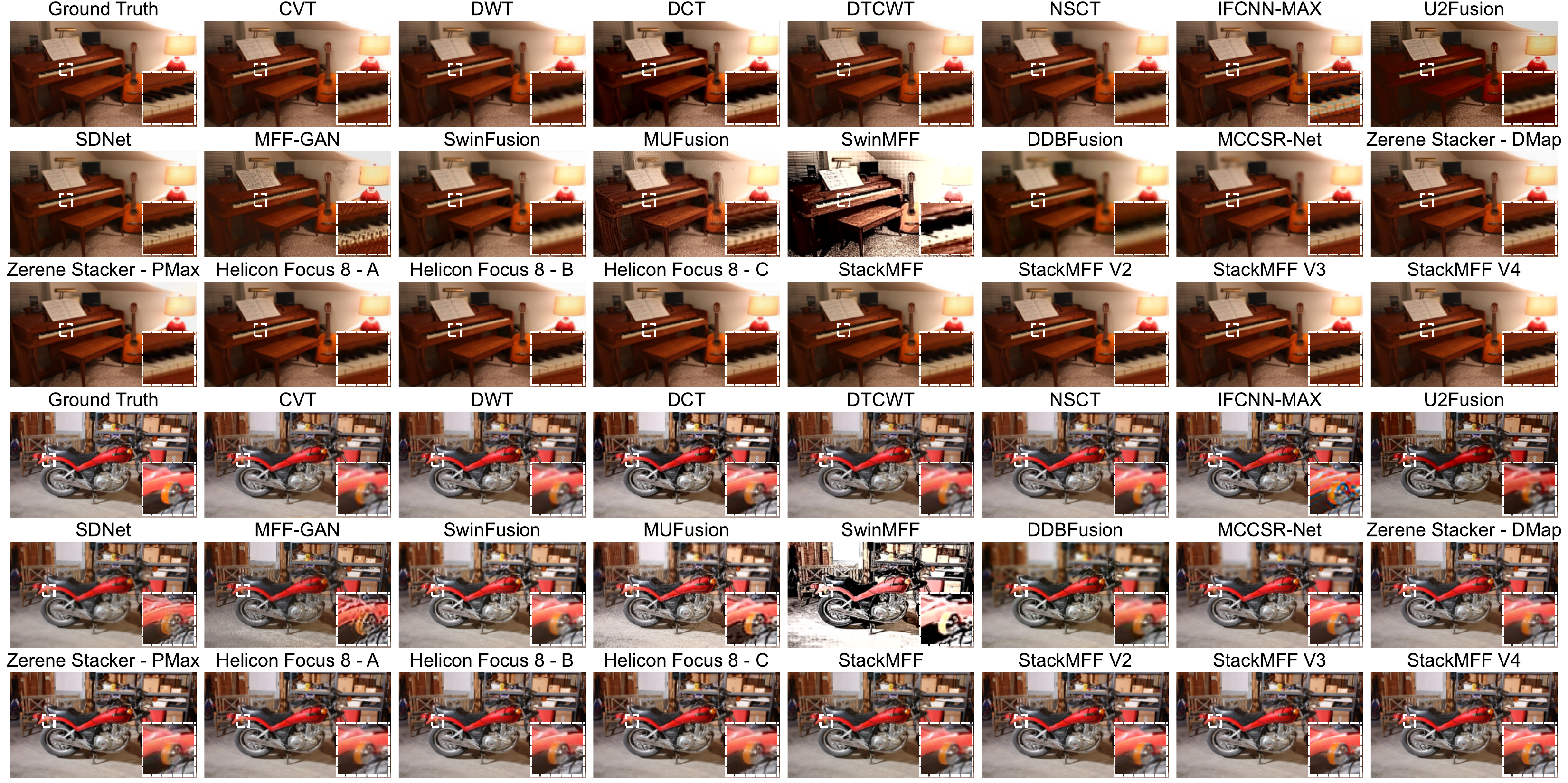}
    \caption{Comparison of fusion results produced by different methods on the Middlebury dataset \cite{scharstein2014high}. The examples correspond to ``piano'' and ``motorcycle'', respectively.}
    \label{fig:stackmffv4_compare_methods_middlebury}
\end{figure*}

For scenes with more intricate details, such as the ``piano'' and ``motorcycle'' examples from the Middlebury dataset (Fig. \ref{fig:stackmffv4_compare_methods_middlebury}), traditional algorithms fail to perform reliably. Specifically, in the ``piano'' example, the magnified regions reveal varying degrees of blurring in the fusion results of CVT, DWT, DTCWT, and NSCT. Pairwise deep learning-based fusion networks are even less effective in these scenarios, exhibiting pronounced noise, visible artifacts, and significant blurring. Among all methods, only commercial focus-stacking software and the StackMFF series yield satisfactory fusion results. In the ``motorcycle'' example, a similar trend is observed: traditional methods exhibit degraded fusion quality in complex scenes, while pairwise fusion networks largely fail. The StackMFF series, particularly StackMFF V4, achieves the most visually compelling fusion results, outperforming those of commercial software.

\subsubsection{Quantitative comparison}
\begin{table*}[htbp]
\small
\setlength{\tabcolsep}{3pt}
\centering
\caption{Quantitative comparison of different multi-focus image fusion methods across four benchmark datasets.}
\begin{tabular}{l cccccccc}
\hline
\textbf{Datasets} & \multicolumn{2}{c}{\textbf{Mobile Depth~\cite{suwajanakorn2015depth}}} & \multicolumn{2}{c}{\textbf{Middlebury~\cite{scharstein2014high}}} & \multicolumn{2}{c}{\textbf{FlyingThings3D~\cite{mayer2016large}}} & \multicolumn{2}{c}{\textbf{Road-MF~\cite{li2024samf}}} \\

\cmidrule(lr){1-1} \cmidrule(lr){2-3} \cmidrule(lr){4-5} \cmidrule(lr){6-7} \cmidrule(lr){8-9}
\textbf{Methods} & \textbf{SSIM} $\uparrow$ & \textbf{PSNR} $\uparrow$ & \textbf{SSIM} $\uparrow$ & \textbf{PSNR} $\uparrow$ & \textbf{SSIM} $\uparrow$ & \textbf{PSNR} $\uparrow$ & \textbf{SSIM} $\uparrow$ & \textbf{PSNR} $\uparrow$ \\
 
% \cmidrule(lr){1-1} \cmidrule(lr){2-2} \cmidrule(lr){3-3} \cmidrule(lr){4-4} \cmidrule(lr){5-5} \cmidrule(lr){6-6} \cmidrule(lr){7-7} \cmidrule(lr){8-8} \cmidrule(lr){9-9}
\cmidrule(lr){1-1} \cmidrule(lr){2-3} \cmidrule(lr){4-5} \cmidrule(lr){6-7} \cmidrule(lr){8-9}
CVT \cite{guo2012multifocus} & 0.9368 (10) & 32.6158 (9) & 0.8893 (11) & 29.3426 (8) & 0.9157 (7) & 30.0917 (9) & 0.9777 (6) & 36.0578 (5) \\
DWT \cite{li1995multisensor} & 0.9340 (11) & 32.1651 (10) & 0.8850 (13) & 29.1761 (10) & 0.9123 (10) & 30.0074 (10) & 0.9309 (11) & 30.3456 (10) \\
DCT \cite{HAGHIGHAT2011789} & 0.4720 (19) & 17.2719 (19) & 0.4520 (19) & 13.9972 (20) & 0.4603 (19) & 15.0949 (19) & 0.4856 (19) & 16.9598 (19) \\
DTCWT \cite{lewis2007pixel} & 0.9412 (7) & 32.7641 (7) & 0.8938 (10) & 29.3763 (7) & 0.9203 (6) & 30.1512 (8) & \cellcolor[gray]{0.9}{0.9826} (2) & 36.7138 (3) \\
NSCT \cite{yang2007image} & 0.9340 (11) & 32.1651 (10) & 0.8850 (13) & 29.1761 (10) & 0.9123 (10) & 30.0074 (10) & 0.9813 (3) & \cellcolor[gray]{0.9}{37.0137} (2) \\
IFCNN-MAX \cite{zhang2020ifcnn} & 0.7882 (17) & 24.9863 (17) & 0.9014 (8) & 29.2064 (9) & 0.9236 (5) & 31.3069 (6) & 0.8952 (15) & 27.6907 (12) \\
U2Fusion \cite{xu2020u2fusion} & 0.3788 (22) & 10.0482 (23) & 0.3980 (23) & 10.1318 (23) & 0.4242 (22) & 11.4382 (24) & 0.3811 (23) & 10.8764 (23) \\
SDNet \cite{zhang2021sdnet} & 0.3961 (21) & 12.1659 (21) & 0.4399 (20) & 14.0048 (19) & 0.4457 (20) & 14.5929 (20) & 0.4144 (21) & 13.0182 (21) \\
MFF-GAN \cite{zhang2021mff} & 0.1797 (24) & 7.1264 (24) & 0.2962 (24) & 10.1180 (24) & 0.3006 (24) & 11.9173 (23) & 0.2559 (24) & 9.3437 (24) \\
SwinFusion \cite{ma2022swinfusion} & 0.4381 (20) & 12.4597 (20) & 0.4254 (21) & 13.4794 (21) & 0.4313 (21) & 14.1286 (21) & 0.3945 (22) & 11.9315 (22) \\
MUFusion \cite{cheng2023mufusion} & 0.4819 (18) & 18.7311 (18) & 0.5809 (18) & 19.7779 (18) & 0.4762 (18) & 19.8073 (18) & 0.6821 (18) & 19.6156 (18) \\
SwinMFF \cite{xie2024swinmff} & 0.3511 (23) & 10.8676 (22) & 0.4215 (22) & 11.8564 (22) & 0.3238 (23) & 12.2809 (22) & 0.4795 (20) & 13.2869 (20) \\
DDBFusion \cite{zhang2025ddbfusion} & 0.8365 (16) & 26.3713 (16) & 0.7181 (17) & 23.7650 (17) & 0.6984 (16) & 23.0223 (17) & 0.8065 (17) & 24.4036 (14) \\
CCSR-Net \cite{ZHENG2025102974} & 0.8485 (15) & 28.3029 (15) & 0.7207 (16) & 24.1580 (16) & 0.6918 (17) & 24.4370 (16) & 0.8682 (16) & 27.5386 (13) \\
MCCSR-Net \cite{ZHENG2025102974} & 0.8750 (14) & 28.5764 (14) & 0.8177 (15) & 26.2944 (12) & 0.7655 (15) & 25.7952 (12) & 0.9090 (14) & 29.9696 (11) \\
Zerene Stacker - DMap & 0.9399 (8) & 33.5643 (5) & 0.9067 (6) & 30.5630 (6) & 0.9139 (9) & 30.9396 (7) & 0.9678 (8) & 33.8450 (7) \\
Zerene Stacker - PMax & 0.9282 (13) & 31.4065 (13) & 0.9068 (5) & 30.6395 (5) & 0.9153 (8) & 31.5620 (5) & 0.9791 (5) & 35.7715 (6) \\
Helicon Focus 8 - A & 0.9469 (5) & 32.9568 (6) & 0.8968 (9) & 24.7029 (15) & 0.8993 (13) & 24.8708 (14) & 0.9203 (13) & 24.1058 (17) \\
Helicon Focus 8 - B \cite{kozub2019focus} & 0.9394 (9) & 33.7037 (4) & 0.8872 (12) & 24.9958 (13) & 0.8965 (14) & 24.8155 (15) & 0.9216 (12) & 24.1157 (16) \\
Helicon Focus 8 - C & 0.9424 (6) & 31.8975 (12) & 0.9028 (7) & 24.8427 (14) & 0.9012 (12) & 25.0471 (13) & 0.9336 (10) & 24.3949 (15) \\
StackMFF \cite{xie2025stackmff} & 0.9536 (3) & 32.6798 (8) & 0.9284 (4) & 31.0764 (4) & 0.9483 (4) & 32.5062 (4) & 0.9692 (7) & 33.0138 (9) \\
StackMFF V2 \cite{xie2025stackmffv2} & 0.9508 (4) & 35.1017 (3) & 0.9444 (3) & 32.1810 (3) & 0.9508 (3) & 32.7506 (3) & 0.9808 (4) & 36.0976 (4) \\
StackMFF V3 & \cellcolor[gray]{0.9}{0.9657} (2) & \cellcolor[gray]{0.9}{36.3498} (2) & \cellcolor[gray]{0.9}{0.9510} (2) & \cellcolor[gray]{0.9}{32.3136} (2) & \cellcolor[gray]{0.9}{0.9607} (2) & \cellcolor[gray]{0.9}{33.3734} (2) & 0.9607 (9) & 33.3734 (8) \\

StackMFF V4 (Ours) & \cellcolor[gray]{0.9}\textbf{0.9733} (1) & \cellcolor[gray]{0.9}\textbf{37.2283} (1) & \cellcolor[gray]{0.9}\textbf{0.9523} (1) & \cellcolor[gray]{0.9}\textbf{32.3604} (1) & \cellcolor[gray]{0.9}\textbf{0.9638} (1) & \cellcolor[gray]{0.9}\textbf{33.7589} (1) & \cellcolor[gray]{0.9}\textbf{0.9938} (1) & \cellcolor[gray]{0.9}\textbf{38.8606} (1) \\
\hline
\end{tabular}
\label{table:compare_methods}
\end{table*}

Table~\ref{table:compare_methods} reports the quantitative fusion performance of the StackMFF V4 method, employed in the deterministic fusion stage, across four benchmark datasets, with higher values indicating better fusion quality. Cells with a gray background but without bold text denote the second-best results, whereas those with both a gray background and bold text denote the best performance.

As shown in Table~\ref{table:compare_methods}, StackMFF V4 consistently achieves the highest SSIM and PSNR across all datasets, demonstrating clear improvements over StackMFF V3 and all other baseline methods. On the Mobile Depth dataset, StackMFF V4 achieves an SSIM of 0.9733 and a PSNR of 37.23 dB, outperforming the second-best method, StackMFF V3, by 0.79\% and 0.88 dB, respectively. Similar trends are observed on the Middlebury, FlyingThings3D, and Road-MF datasets.

In comparison, traditional transform-based methods, such as DTCWT and NSCT, exhibit relatively stable performance across datasets, indicating that manually designed rules remain highly effective for image stack fusion tasks. However, many learning-based methods originally designed for image-pair fusion, including IFCNN, U2Fusion, SDNet, MFF-GAN, and SwinFusion, perform poorly on image stack fusion due to error accumulation when applied sequentially to multi-layer stacks.

Commercial focus-stacking software, such as Helicon Focus 8 and Zerene Stacker, generally relies on traditional fusion strategies and achieves competitive performance. Overall, the results in Table~\ref{table:compare_methods} clearly illustrate that StackMFF V4 not only quantitatively surpasses all baseline methods but also establishes a new benchmark for deterministic multi-focus image stack fusion.

\subsubsection{Model efficiency comparison}
\begin{table}
    \centering
    \small
    \setlength{\tabcolsep}{3pt}
    \caption{Comparison of learning-based methods in terms of model size (M), computational cost (FLOPs, in G) for fusing two $256\times256$ images, and average runtime (s) on the Mobile Depth dataset~\cite{suwajanakorn2015depth}.}
    \begin{tabular}{l c c c}
    \hline
        \textbf{Methods} & \textbf{Model Size (M)} & \textbf{FLOPs (G)} & \textbf{Time (s)} \\ \hline
        CVT~\cite{guo2012multifocus} & - & - & 48.00 \\
        DWT~\cite{li1995multisensor} & - & - & 5.34 \\ 
        DCT~\cite{HAGHIGHAT2011789} & - & - & 4.97 \\ 
        DTCWT~\cite{lewis2007pixel} & - & - & 11.44 \\ 
        NSCT~\cite{yang2007image} & - & - & 231.84 \\ 
        IFCNN-MAX \cite{zhang2020ifcnn} & 0.08 & 8.54 & 0.55 \\
        U2Fusion \cite{xu2020u2fusion} & 0.66 & 86.4 & 41.04 \\
        SDNet \cite{zhang2021sdnet} & 0.07 & 8.81 & 9.68 \\
        MFF-GAN \cite{zhang2021mff} & 0.05 & 3.08 & 6.40 \\
        SwinFusion \cite{ma2022swinfusion} & 0.93 & 63.73 & 28.21 \\
        MUFusion \cite{cheng2023mufusion} & 2.16 & 24.07 & 40.40 \\
        SwinMFF \cite{xie2024swinmff} & 41.25 & 22.38 & 27.97 \\
        DDBFusion \cite{zhang2025ddbfusion} & 10.92 & 184.93 & 33.89 \\
        CCSR-Net \cite{ZHENG2025102974} & \cellcolor[gray]{0.9}\textbf{0.02} & \cellcolor[gray]{0.9}1.59 & 1.92 \\
        MCCSR-Net \cite{ZHENG2025102974} & \cellcolor[gray]{0.9}{0.04} & 2.57 & 16.03 \\
        Zerene Stacker - DMap & - & - & 14.64 \\
        Zerene Stacker - PMax & - & - & 6.36 \\
        Helicon Focus 8 - A & - & - & 0.30 \\
        Helicon Focus 8 - B \cite{kozub2019focus} & - & - & 0.38 \\
        Helicon Focus 8 - C & - & - & 0.33 \\
        StackMFF \cite{xie2025stackmff} & 6.08 & 21.98 & 0.22 \\
        StackMFF V2 \cite{xie2025stackmffv2} & 0.05 & 2.75 & \cellcolor[gray]{0.9}0.14 \\
        StackMFF V3 & 2.74 & 2.04 & 0.52 \\
        StackMFF V4 (Ours) & 0.94 & \cellcolor[gray]{0.9}\textbf{0.51} & \cellcolor[gray]{0.9}\textbf{0.12} \\
    \hline
    \end{tabular}
    \label{table:efficiency_stackmffv4}
\end{table}

Table~\ref{table:efficiency_stackmffv4} presents a comprehensive comparison of various MFF methods in terms of efficiency, including model size, computational cost, and runtime performance. The results clearly demonstrate that StackMFF V4 achieves outstanding computational efficiency while maintaining superior fusion quality, requiring only 0.51G FLOPs to process two $256\times256$ images, significantly outperforming competing methods. Notably, on the Mobile Depth dataset, the average processing time per image stack for StackMFF V4 is only 0.12 seconds—approximately four times faster than StackMFF V3—making it the fastest method and a practical solution for real-time multi-focus image fusion. In contrast, many existing approaches, such as DDBFusion and U2Fusion, demand excessive computational resources and long inference times, while several methods, including most image-pair fusion networks like MFF-GAN and SwinMFF, exhibit insufficient fusion quality for image stack fusion tasks. These results underscore the dual advantages of our method in terms of both efficiency and fusion quality. Therefore, StackMFF V4 can provide reliable conditional images for the generative restoration stage, serving as input to generate high-fidelity images.

\subsubsection{Ablation studies}
In this subsection, we present an ablation study to demonstrate the three key improvements of StackMFF V4 over StackMFF V3. All ablation experiments are conducted on the Mobile Depth dataset~\cite{suwajanakorn2015depth}.

\begin{table}[!ht]
    \centering
    \small
    \setlength{\tabcolsep}{2pt}
    \caption{Comparison of intra-layer focus estimation networks across different StackMFF series versions.}
    \begin{tabular}{lcccc}
        \toprule
        \textbf{Methods} & \textbf{SSIM $\uparrow$} & \textbf{PSNR $\uparrow$} & \textbf{Model Size (M)} & \textbf{FLOPs (G)} \\
        \midrule
        ULDA-Net (V2) & 0.9716 & 37.0408 & \cellcolor[gray]{0.9}\textbf{0.03} & \cellcolor[gray]{0.9}0.59 \\
        PFMLP (V3) & \cellcolor[gray]{0.9}0.9718 & \cellcolor[gray]{0.9}37.1114 & 2.75 & 1.79 \\
        MobilieNetV3 (V4) & \cellcolor[gray]{0.9}\textbf{0.9733} & \cellcolor[gray]{0.9}\textbf{37.2283} & \cellcolor[gray]{0.9}0.94 & \cellcolor[gray]{0.9}\textbf{0.51} \\
        \bottomrule
    \end{tabular}
    \label{table:ablation_network}
\end{table}

First, Table~\ref{table:ablation_network} compares the impact of different intra-layer focus estimation networks on the fusion results. StackMFF V2 employs the custom-designed ULDA-Net, StackMFF V3 utilizes a modified variant based on PFMLP~\cite{huangpfmlp}, and StackMFF V4 implements a network based on MobileNetV3-Large~\cite{howard2019searching}. To ensure a fair comparison, the cross-layer interaction module is fixed as the SACA introduced in StackMFF V4.

\begin{table}[!ht]
    \centering
    \small
    \setlength{\tabcolsep}{4pt}
    \caption{Comparison of attention modules used in StackMFF V3 (Pixel-wise Cross-Layer Attention, PCA) and StackMFF V4 (Spatial Aggregation Cross-Layer Attention, SACA).}
    \begin{tabular}{lcccc}
        \toprule
        \textbf{Attention} & \textbf{SSIM $\uparrow$} & \textbf{PSNR $\uparrow$} & \textbf{Model Size (M)} & \textbf{FLOPs (G)} \\
        \midrule
        PCA (V3) & \cellcolor[gray]{0.9}0.9728 & \cellcolor[gray]{0.9}37.1479 & \cellcolor[gray]{0.9}0.94 & \cellcolor[gray]{0.9}0.93 \\
        SACA (V4) & \cellcolor[gray]{0.9}\textbf{0.9733} & \cellcolor[gray]{0.9}\textbf{37.2283} & \cellcolor[gray]{0.9}\textbf{0.94} & \cellcolor[gray]{0.9}\textbf{0.51} \\
        \bottomrule
    \end{tabular}
    \label{table:ablation_attention}
\end{table}

Another key improvement of StackMFF V4 over StackMFF V3 lies in its attention module for cross-layer interactions. In this ablation study, we specifically compare the attention modules used for cross-layer interactions while keeping all other network components unchanged. In particular, the intra-layer focus estimation network is consistently implemented using MobileNetV3-Large, as in StackMFF V4. StackMFF V4 introduces the Spatial Aggregation Cross-Layer Attention (SACA), which, compared to the Pixel-wise Cross-Layer Attention (PCA) used in StackMFF V3, not only significantly improves fusion performance but also substantially reduces computational cost, thereby providing dual benefits. This improvement is quantitatively illustrated in Table~\ref{table:ablation_attention}.

\begin{table}[!ht]
    \centering
    \small
    \setlength{\tabcolsep}{3pt}
    \caption{Effect of the number of iterative refinement loops on the fusion performance of StackMFF V4.}
    \begin{tabular}{ccccc}
        \toprule
        \textbf{Num. of Loops} & \textbf{SSIM $\uparrow$} & \textbf{PSNR $\uparrow$} & \textbf{Model Size (M)} & \textbf{FLOPs (G)} \\
        \midrule
        0 & 0.9688 & 36.4529 & \cellcolor[gray]{0.9}\textbf{0.94} & \cellcolor[gray]{0.9}\textbf{0.41} \\
        1 & \cellcolor[gray]{0.9}0.9733 & \cellcolor[gray]{0.9}37.2283 & \cellcolor[gray]{0.9}0.94 & \cellcolor[gray]{0.9}0.51 \\
        2 & \cellcolor[gray]{0.9}\textbf{0.9754} & \cellcolor[gray]{0.9}\textbf{37.5507} & 0.94 & 0.60 \\
        4 & 0.9729 & 36.8328 & 0.95 & 0.79 \\
        8 & 0.9715 & 36.8203 & 0.96 & 1.17 \\
        \bottomrule
    \end{tabular}
    \label{table:ablation_loop}
\end{table}

We further conduct an ablation study on the number of iterative refinement loops in StackMFF, as shown in Table~\ref{table:ablation_loop}. Increasing the number of loops from 0 to 1, i.e., introducing a single iterative refinement, leads to a substantial improvement in fusion performance. Increasing the number of loops from 1 to 2 yields only a marginal gain. However, further increasing the number of loops to 4 or 8 results in a slight degradation in performance. Considering that the additional gain from increasing the number of loops beyond 1 is not cost-effective given the additional computational overhead, StackMFF V4 adopts a single iterative refinement loop.

\begin{table}[!ht]
    \centering
    \small
    \setlength{\tabcolsep}{16pt}
    \caption{Effect of different spatial aggregation downsampling ratios in SACA on fusion performance.}
    \begin{tabular}{ccc}
        \toprule
        \textbf{Downsampling Ratio} & \textbf{SSIM $\uparrow$} & \textbf{PSNR $\uparrow$} \\
        \midrule
        $1/2$ & \cellcolor[gray]{0.9}0.9732 & 37.0580 \\
        $1/4$ & \cellcolor[gray]{0.9}\textbf{0.9733} & \cellcolor[gray]{0.9}\textbf{37.2283} \\
        $1/8$ & \cellcolor[gray]{0.9}0.9732 & 37.0069 \\
        $1/16$ & 0.9717 & \cellcolor[gray]{0.9}37.2135 \\
        \bottomrule
    \end{tabular}
    \label{table:ablation_saca_downsample}
\end{table}

To determine the optimal aggregation range for SACA, we further investigate the impact of different downsampling ratios on fusion performance, as shown in Table~\ref{table:ablation_saca_downsample}. A downsampling ratio of $1/2$ indicates that both the height and width are reduced to half of their original dimensions, with other ratios defined analogously. The quantitative results show that fusion quality initially improves and then declines as the aggregation range expands. Notably, when the downsampling ratio is set to $1/4$, both evaluation metrics reach their peak values. Therefore, StackMFF V4 adopts a downsampling ratio of $1/4$, corresponding to aggregating each $4 \times 4$ region into a single patch for cross-layer attention computation.

\subsection{Evaluation of GMFF}
\subsubsection{Qualitative comparison}

\begin{figure*} % 
    \centering
    \includegraphics[width=\linewidth]{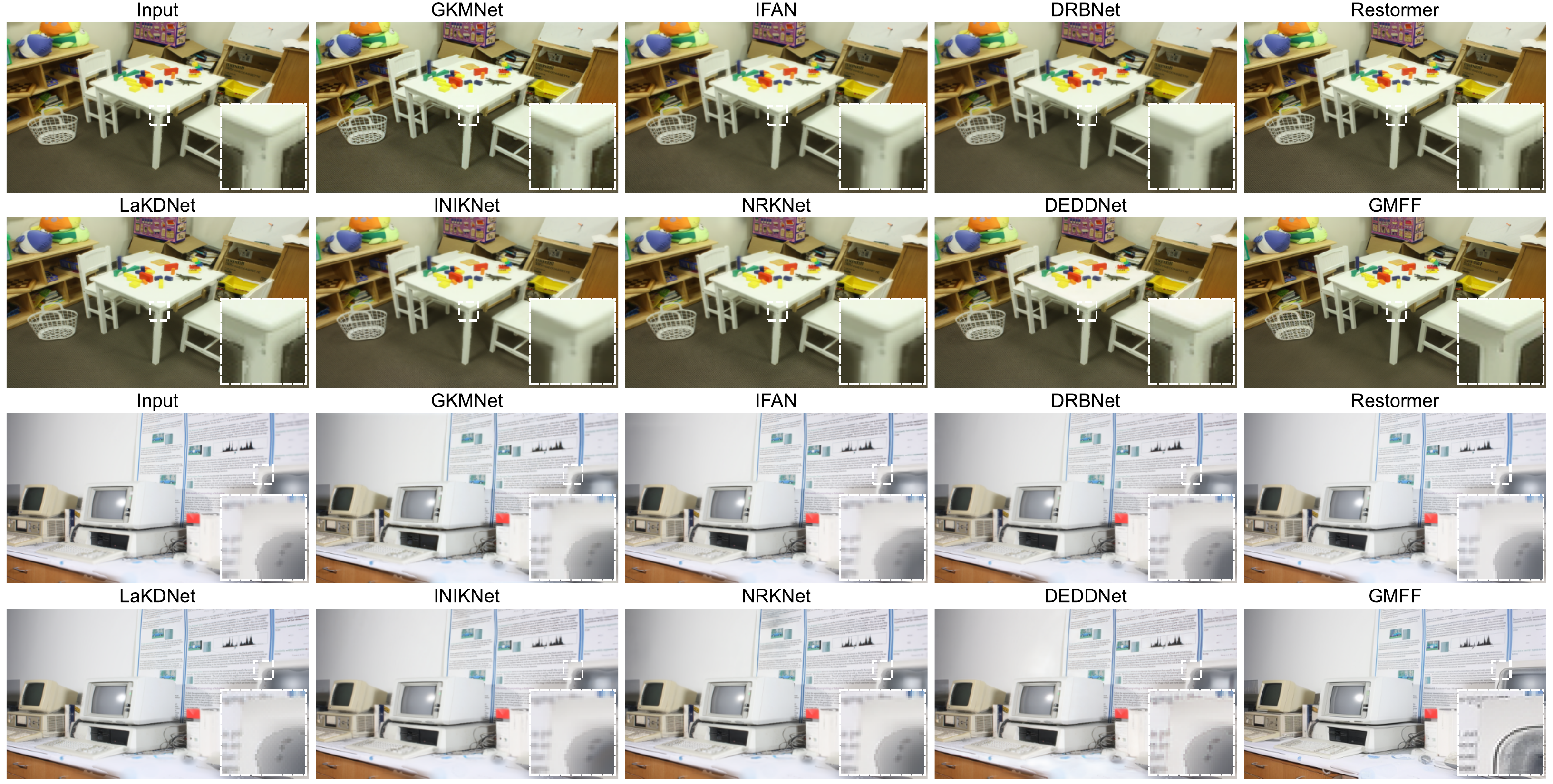}
    \caption{Comparison of the results produced by different methods on the Middlebury dataset \cite{scharstein2014high}. The examples correspond to ``Playtable'' and ``Vintage,'' respectively. All models take as input the fused output of the StackMFF V4 network.}
    \label{fig:gmff_compare_methods_middlebury}
\end{figure*}

We present the final fused results of the proposed GMFF framework in Fig.~\ref{fig:gmff_compare_methods_middlebury} and compare them with the outputs of other image restoration models that are specifically designed for defocus deblurring. All models take as input the fused output of the StackMFF V4 network, which is used in the deterministic fusion stage of the GMFF framework.

The first example, ``Playtable,'' illustrates the primary function of the generative restoration stage—edge artifact removal. From the enlarged regions, it can be observed that when the input fused image still contains noticeable edge artifacts that degrade visual quality, the generative restoration stage effectively suppresses these artifacts and reconstructs realistic edge details, thereby enhancing the overall fusion quality.

The second example, ``Vintage,'' demonstrates another key function of the generative restoration stage—completion of missing focal plane information. Whether the blurring arises from failures in the deterministic fusion stage or from missing focal plane data in the original inputs, the generative restoration stage produces plausible restored content conditioned on the output of the first-stage fusion. This indicates that the generative restoration stage is robust to variations in the quality of the initial fused inputs.

Comparisons with other image restoration models further highlight their limitations. Although these models are specifically trained for defocus deblurring, the results for the ``Playtable'' example show that they are not effective at removing edge artifacts or correcting edge blurring. Similarly, in the ``Vintage'' example, where most regions are already clear and only a small area is defocused, these deblurring methods largely fail. Only the proposed GMFF is capable of regenerating missing details through generative restoration.

\begin{figure*} % 
    \centering
    \includegraphics[width=\linewidth]{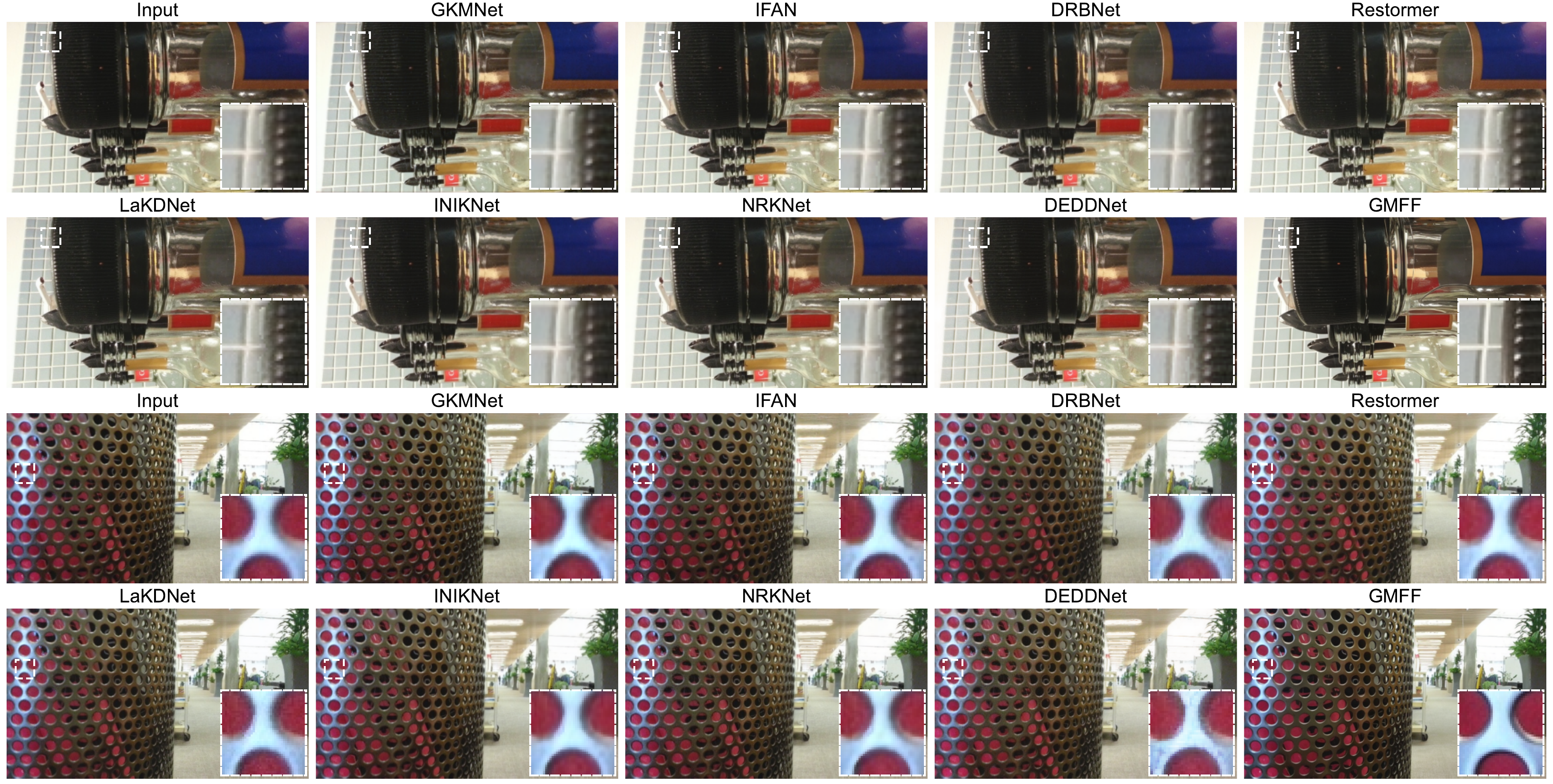}
    \caption{Comparison of the results produced by different methods on the Mobile Depth dataset \cite{suwajanakorn2015depth}. The examples correspond to ``Bottles'' and ``Bucket,'' respectively. All models take as input the fused output of the StackMFF V4 network.}
    \label{fig:gmff_compare_methods_mobile_depth}
\end{figure*}

Fig.~\ref{fig:gmff_compare_methods_mobile_depth} illustrates the third key function of the generative restoration stage—detail refinement and image quality enhancement. In the ``Bottles'' example, it can be observed that GMFF further enhances fine image details. Specifically, the serrated textures on the bottle caps become more pronounced, the transitions between the bottle caps and the background wall edges appear sharper, and the tile patterns on the background wall become more clearly visible.

The ``Bucket'' example further demonstrates this advantage. When the input image is of relatively low quality and exhibits noticeable blurring, GMFF can generate visually plausible details conditioned on the input, the learned generative priors, and the provided guidance. This results in a substantial improvement in the overall image quality.

\begin{figure*} 
    \centering
    \includegraphics[width=\linewidth]{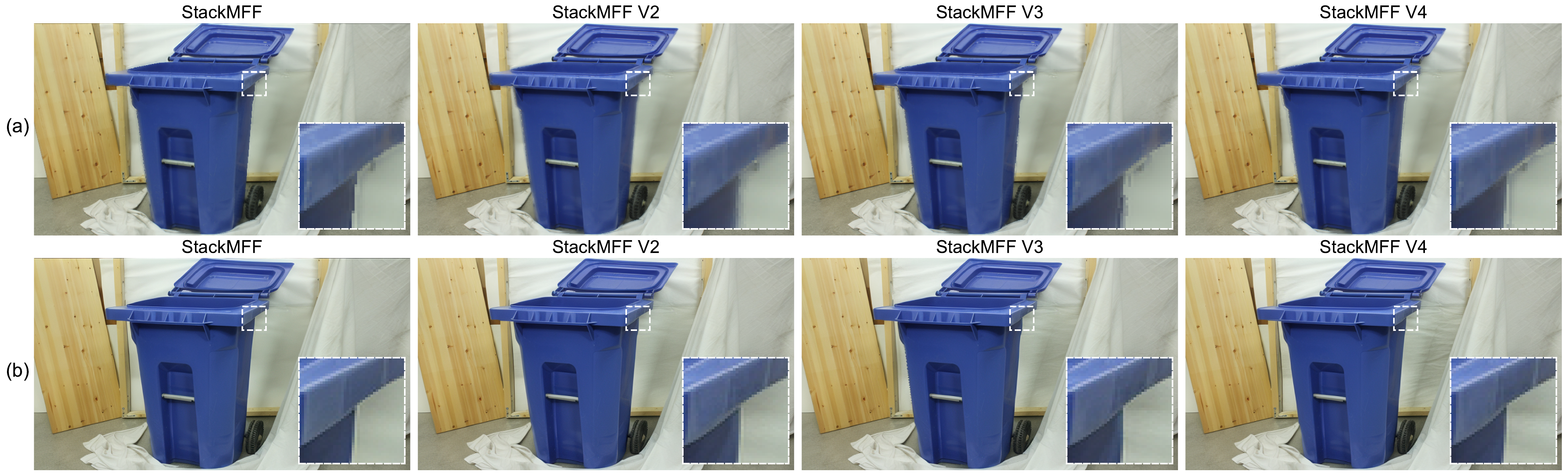}
    \caption{Comparison of the results before and after the generative restoration stage, where the fused outputs from different deterministic fusion models in the StackMFF series serve as conditional input images. (a) Conditional images, i.e., the fused outputs from the StackMFF series; (b) Final images obtained after the generative restoration stage.}

    \label{fig:comparison_gmff_before_and_after_Recycle}
\end{figure*}

To further demonstrate that the proposed generative restoration stage is decoupled from deterministic fusion models and can effectively address edge artifacts—a long-standing challenge in the MFF domain—Fig.~\ref{fig:comparison_gmff_before_and_after_Recycle} presents a comparison of the results before and after the generative restoration stage. In this comparison, the fused outputs from different deterministic fusion models in the StackMFF series serve as conditional input images.

Although the StackMFF series represents the state of the art in multi-focus image stack fusion, it still fails to completely eliminate edge artifacts. In contrast, the proposed generative restoration stage effectively mitigates this issue by leveraging learned generative priors—a capability that previous fusion models, particularly those based on decision maps, find difficult to achieve.

\subsubsection{Quantitative comparison}

\begin{table}[htbp]
\small
\setlength{\tabcolsep}{2pt}
\centering
\caption{Quantitative evaluation results on the Mobile Depth and Middlebury datasets. All models take as input the fused output of the StackMFF V4 network.}
\begin{tabular}{lcccc}
\hline
\textbf{Datasets} & \multicolumn{2}{c}{\textbf{Mobile Depth~\cite{suwajanakorn2015depth}}} & \multicolumn{2}{c}{\textbf{Middlebury~\cite{scharstein2014high}}} \\
\cmidrule(lr){1-1} \cmidrule(lr){2-3} \cmidrule(lr){4-5}
\textbf{Methods} & \textbf{BRISQUE} $\downarrow$ & \textbf{PIQE} $\downarrow$ & \textbf{BRISQUE} $\downarrow$ & \textbf{PIQE} $\downarrow$ \\
\cmidrule(lr){1-1} \cmidrule(lr){2-2} \cmidrule(lr){3-3} \cmidrule(lr){4-4} \cmidrule(lr){5-5}

Input & 14.98 (8) & 28.00 (4) & 25.87 (4) & 44.28 (5) \\
GKMNet~\cite{quan2021gaussian} & 15.21 (9) & 33.09 (9) & 29.92 (8) & 42.76 (4) \\
IFAN~\cite{lee2021iterative} & 14.53 (6) & 30.28 (6) & 29.08 (6) & 51.79 (9) \\
DRBNet~\cite{ruan2022learning} & 14.90 (7) & 31.78 (7) & 25.25 (3) & 45.99 (6) \\
Restormer~\cite{zamir2022restormer} & 12.60 (4) & 29.76 (5) & 34.33 (9) & 49.70 (7) \\
LaKDNet~\cite{ruan2023revisiting} & 12.48 (3) & \cellcolor[gray]{0.9}\textbf{18.70 (1)} & 28.37 (5) & \cellcolor[gray]{0.9}\textbf{24.12 (1)} \\
INIKNet~\cite{quan2023single} & 17.50 (10) & 38.90 (10) & 29.52 (7) & 53.04 (10) \\
NRKNet~\cite{quan2023neumann} & 14.36 (5) & 32.39 (8) & 36.80 (10) & 51.65 (8) \\
DEDDNet~\cite{zhai2024efficient} & \cellcolor[gray]{0.9}11.34 (2) & \cellcolor[gray]{0.9}24.65 (2) & \cellcolor[gray]{0.9}23.39 (2) & 39.23 (3) \\
GMFF (Ours) & \cellcolor[gray]{0.9}\textbf{9.20 (1)} & 27.25 (3) & \cellcolor[gray]{0.9}\textbf{13.67 (1)} & \cellcolor[gray]{0.9}29.35 (2) \\
\hline
\end{tabular}
\label{table:brisque_piqe_results}
\end{table}

We quantitatively compare the performance of different image restoration models in Table~\ref{table:brisque_piqe_results}, and also provide the evaluation results of the input images for reference. As shown in the table, the proposed GMFF achieves the lowest BRISQUE scores on both the Mobile Depth and Middlebury datasets, indicating that the images restored through the generative fusion process exhibit enhanced naturalness and clarity, and align more closely with the characteristics of high-quality natural images.

Moreover, compared with the StackMFF V4 inputs, both BRISQUE and PIQE scores consistently decrease across datasets, demonstrating the effectiveness of the proposed generative restoration stage in enhancing perceptual quality. On the Middlebury dataset, GMFF achieves the second-lowest PIQE score, suggesting that in complex scenes with abundant fine details, the proposed method effectively suppresses regional artifacts and distortions while preserving structural fidelity.

In contrast, existing image restoration models—even those specifically trained for defocus deblurring—generally fail to further enhance the fused outputs produced by multi-focus image fusion algorithms. In some cases, minor degradations are observed; for instance, GKMNet and INIKNet slightly increase the BRISQUE scores, showing similar trends for PIQE. This can be attributed to the fact that, after fusion, most regions are already near-optimal in clarity, while only certain complex areas—such as edges or textured regions—may still contain residual fusion artifacts or localized blur. Existing deblurring networks are not well adapted to handle such localized imperfections, which explains their limited effectiveness in this context.

\begin{table*}[htbp]
\small
\setlength{\tabcolsep}{8pt}
\centering
\caption{Quantitative comparison of the different MFF methods before and after the proposed generative restoration stage.}
\begin{tabular}{l cccc}
\hline
\textbf{Datasets} & \multicolumn{2}{c}{\textbf{Mobile Depth~\cite{suwajanakorn2015depth}}} & \multicolumn{2}{c}{\textbf{Middlebury~\cite{scharstein2014high}}} \\
\cmidrule(lr){1-1} \cmidrule(lr){2-3} \cmidrule(lr){4-5}

\textbf{Methods} & \multicolumn{1}{c}{\textbf{BRISQUE} $\downarrow$} & \multicolumn{1}{c}{\textbf{PIQE} $\downarrow$} & \multicolumn{1}{c}{\textbf{BRISQUE} $\downarrow$} & \multicolumn{1}{c}{\textbf{PIQE} $\downarrow$} \\
\cmidrule(lr){1-1} \cmidrule(lr){2-2} \cmidrule(lr){3-3} \cmidrule(lr){4-4} \cmidrule(lr){5-5}

CVT \cite{guo2012multifocus} & 22.80 (10)$\rightarrow$16.88 (9) & 39.17 (9)$\rightarrow$33.57 (6) & 35.31 (14)$\rightarrow$13.82 (9) & 47.42 (12)$\rightarrow$33.59 (14) \\
DWT \cite{li1995multisensor} & 23.66 (11)$\rightarrow$16.19 (6) & 39.99 (11)$\rightarrow$35.18 (13) & 33.37 (12)$\rightarrow$14.05 (12) & 47.23 (7)$\rightarrow$33.74 (15) \\
DCT \cite{HAGHIGHAT2011789} & 23.82 (13)$\rightarrow$18.92 (17) & 39.80 (10)$\rightarrow$38.93 (19) & 31.71 (10)$\rightarrow$15.49 (19) & 47.00 (6)$\rightarrow$36.50 (20) \\
DTCWT \cite{lewis2007pixel} & 23.83 (14)$\rightarrow$17.27 (11) & 41.09 (14)$\rightarrow$33.90 (9) & 35.51 (15)$\rightarrow$14.30 (15) & 47.37 (10)$\rightarrow$34.30 (18) \\
NSCT \cite{yang2007image} & 23.66 (11)$\rightarrow$16.19 (6) & 39.99 (11)$\rightarrow$35.18 (13) & 33.37 (12)$\rightarrow$14.05 (12) & 47.23 (7)$\rightarrow$33.74 (15) \\
IFCNN-MAX \cite{zhang2020ifcnn} & \cellcolor{red!25}10.82 (1)$\rightarrow$18.37 (15) & \cellcolor{red!25}26.16 (2)$\rightarrow$30.44 (2) & \cellcolor[gray]{0.9}19.47 (1)$\rightarrow$8.14 (2) & \cellcolor[gray]{0.9}\textbf{31.23 (1)$\rightarrow$26.11 (1)} \\
U2Fusion \cite{xu2020u2fusion} & 28.53 (19)$\rightarrow$23.36 (21) & 44.32 (18)$\rightarrow$37.37 (17) & 37.33 (18)$\rightarrow$17.11 (21) & 54.97 (19)$\rightarrow$28.96 (4) \\
SDNet \cite{zhang2021sdnet} & 16.50 (4)$\rightarrow$16.17 (5) & \cellcolor{red!25}25.96 (1)$\rightarrow$32.30 (3) & \cellcolor[gray]{0.9}\textbf{38.04 (20)$\rightarrow$5.91 (1)} & 48.30 (14)$\rightarrow$28.83 (3) \\
MFF-GAN \cite{zhang2021mff} & 93.39 (24)$\rightarrow$41.55 (23) & 58.37 (23)$\rightarrow$34.97 (11) & 73.00 (24)$\rightarrow$58.79 (24) & 49.43 (15)$\rightarrow$45.88 (23) \\
SwinFusion \cite{ma2022swinfusion} & \cellcolor[gray]{0.9}19.84 (8)$\rightarrow$14.67 (2) & 42.10 (15)$\rightarrow$33.65 (7) & 35.96 (16)$\rightarrow$13.90 (11) & 57.86 (22)$\rightarrow$31.10 (8) \\
MUFusion \cite{cheng2023mufusion} & 34.25 (22)$\rightarrow$31.55 (22) & 50.91 (20)$\rightarrow$38.52 (18) & 36.17 (17)$\rightarrow$16.02 (20) & 49.76 (16)$\rightarrow$43.32 (22) \\
SwinMFF \cite{xie2024swinmff} & 62.53 (23)$\rightarrow$52.69 (24) & 65.98 (24)$\rightarrow$56.29 (24) & 65.25 (23)$\rightarrow$45.45 (23) & 64.31 (24)$\rightarrow$59.83 (24) \\
DDBFusion \cite{zhang2025ddbfusion} & 32.95 (21)$\rightarrow$17.84 (14) & 51.26 (21)$\rightarrow$39.81 (20) & 37.75 (19)$\rightarrow$20.03 (22) & 56.54 (21)$\rightarrow$41.85 (21) \\
CCSR-Net \cite{ZHENG2025102974} & 21.69 (9)$\rightarrow$18.46 (16) & 38.47 (8)$\rightarrow$35.11 (12) & 59.89 (22)$\rightarrow$14.20 (14) & 61.80 (23)$\rightarrow$31.91 (9) \\
MCCSR-Net \cite{ZHENG2025102974} & 17.55 (5)$\rightarrow$17.28 (12) & \cellcolor{red!25}30.08 (5)$\rightarrow$36.40 (16) & 39.31 (21)$\rightarrow$12.37 (4) & \cellcolor[gray]{0.9}54.06 (17)$\rightarrow$28.50 (2) \\
Zerene Stacker - DMap & 27.12 (18)$\rightarrow$15.72 (4) & 43.66 (17)$\rightarrow$34.60 (10) & 31.23 (9)$\rightarrow$14.82 (18) & 45.03 (4)$\rightarrow$34.30 (17) \\
Zerene Stacker - PMax & 25.03 (15)$\rightarrow$16.64 (8) & 40.76 (13)$\rightarrow$36.13 (15) & 29.20 (7)$\rightarrow$13.79 (7) & 45.06 (5)$\rightarrow$32.64 (11) \\
Helicon Focus 8 - A & 29.93 (20)$\rightarrow$19.98 (18) & 52.72 (22)$\rightarrow$42.70 (22) & 32.94 (11)$\rightarrow$14.57 (17) & 54.75 (18)$\rightarrow$35.31 (19) \\
Helicon Focus 8 - B \cite{kozub2019focus} & \cellcolor{red!25}19.79 (7)$\rightarrow$20.02 (19) & 34.67 (7)$\rightarrow$41.52 (21) & 28.69 (6)$\rightarrow$13.82 (8) & 47.94 (13)$\rightarrow$31.96 (10) \\
Helicon Focus 8 - C & 27.05 (17)$\rightarrow$21.69 (20) & 45.84 (19)$\rightarrow$42.87 (23) & 31.18 (8)$\rightarrow$14.33 (16) & 55.20 (20)$\rightarrow$32.74 (12) \\
StackMFF \cite{xie2025stackmff} & 26.24 (16)$\rightarrow$14.95 (3) & 42.57 (16)$\rightarrow$33.78 (8) & 27.04 (4)$\rightarrow$12.29 (3) & 47.39 (11)$\rightarrow$32.99 (13) \\
StackMFF V2 \cite{xie2025stackmffv2} & 19.25 (6)$\rightarrow$17.44 (13) & \cellcolor{red!25}30.87 (6)$\rightarrow$33.50 (5) & 27.79 (5)$\rightarrow$13.65 (5) & 47.29 (9)$\rightarrow$30.85 (7) \\
StackMFF V3 & \cellcolor{red!25}14.19 (2)$\rightarrow$17.05 (10) & 27.01 (3)$\rightarrow$32.89 (4) & 25.75 (2)$\rightarrow$13.88 (10) & 44.56 (3)$\rightarrow$30.22 (6) \\
StackMFF V4 (Ours) & \cellcolor[gray]{0.9}\textbf{15.07 (3)$\rightarrow$9.20 (1)} & \cellcolor[gray]{0.9}\textbf{29.54 (4)$\rightarrow$27.25 (1)} & 25.87 (3)$\rightarrow$13.67 (6) & 44.28 (2)$\rightarrow$29.35 (5) \\
\hline
\end{tabular}
\label{table:compare_inputs_gmff}
\end{table*}

As shown in Table~\ref{table:compare_inputs_gmff}, the table presents the quantitative differences before and after the generative restoration stage when the fused images from different multi-focus image fusion methods are used as conditional inputs. The BRISQUE and PIQE scores before and after restoration are connected by arrows ($\rightarrow$), indicating changes in perceptual quality. Cells with red backgrounds denote rare cases in which the restoration results in performance degradation, suggesting potential adverse optimization effects.

The results indicate that the proposed generative restoration stage is compatible with most deterministic fusion methods and consistently produces perceptual improvements in the majority of cases. This demonstrates the generality and robustness of the restoration framework in enhancing image quality beyond that of conventional fusion outputs. Moreover, when the conditional image is provided by our proposed StackMFF~V4, the method achieves the best performance in both BRISQUE and PIQE metrics on the Mobile Depth dataset. The restoration effect remains consistently positive, further corroborating the effectiveness of our approach.

\subsubsection{Model efficiency comparison}

\begin{table}[htbp]
\centering
\small
\setlength{\tabcolsep}{2pt}
\caption{The efficiency comparison of the different methods on the Mobile Depth~\cite{suwajanakorn2015depth} and Middlebury~\cite{scharstein2014high} datasets, with FLOPs (in billions, G), inference time (in seconds, s), and model size (in millions of parameters, M).}
\begin{tabular}{lccccc}
\toprule
\multirow{2}{*}{\textbf{Methods}} & \multicolumn{2}{c}{\textbf{Mobile Depth~\cite{suwajanakorn2015depth}}} & \multicolumn{2}{c}{\textbf{Middlebury~\cite{scharstein2014high}}} & \multirow{2}{*}{\textbf{Model Size}} \\
\cmidrule(lr){2-3} \cmidrule(lr){4-5} 
 & \textbf{FLOPs} & \textbf{Time} & \textbf{FLOPs} & \textbf{Time} & \\
\cmidrule(lr){1-1} \cmidrule(lr){2-2} \cmidrule(lr){3-3} \cmidrule(lr){4-4} \cmidrule(lr){5-5} \cmidrule(lr){6-6} 
GKMNet~\cite{quan2021gaussian}    & \cellcolor[gray]{0.9}\textbf{75.66}  & 0.07  & \cellcolor[gray]{0.9}\textbf{114.71} & 0.10  & \cellcolor[gray]{0.9}\textbf{1.41}    \\
IFAN~\cite{lee2021iterative}      & \cellcolor[gray]{0.9}104.93 & \cellcolor[gray]{0.9}\textbf{0.02}  & \cellcolor[gray]{0.9}160.83 & \cellcolor[gray]{0.9}\textbf{0.03}  & 10.48   \\
DRBNet~\cite{ruan2022learning}    & 169.46 & \cellcolor[gray]{0.9}0.05  & 262.66 & \cellcolor[gray]{0.9}0.04  & 44.59   \\
Restormer~\cite{zamir2022restormer} & 544.51 & 0.35  & 843.99 & 0.43  & 26.13   \\
LaKDNet~\cite{ruan2023revisiting}   & 342.66 & 0.29  & 419.96 & 0.36  & 17.73   \\
INIKNet~\cite{quan2023single}   & 272.39 & 0.17  & 408.55 & 0.18  & \cellcolor[gray]{0.9}1.98    \\
NRKNet~\cite{quan2023neumann}    & 321.47 & 0.06  & 392.05 & 0.10  & 6.09    \\
DEDDNet~\cite{zhai2024efficient}   & 274.06 & 0.11  & 420.22 & 0.09  & 4.69    \\
GMFF (Ours)      & 492.95 & 17.63 & 603.70 & 13.62 & 6358.15 \\
\bottomrule
\end{tabular}
\label{table:imgae_restoration_efficiency_comparison}
\end{table}
Table~\ref{table:imgae_restoration_efficiency_comparison} reports the efficiency comparison of different image restoration methods on the Mobile Depth~\cite{suwajanakorn2015depth} and Middlebury~\cite{scharstein2014high} datasets. As shown, our proposed GMFF method has a substantially larger model size and slower inference speed compared with existing state-of-the-art image restoration networks. This is primarily due to GMFF leveraging a Stable Diffusion 2.1-base backbone, benefiting from the richer priors provided by its larger model capacity and large-scale pretraining, and the multi-step sampling process inherent to diffusion models, which increases inference time. Despite these computational costs, GMFF achieves clear advantages in enhancing the quality of fused multi-focus images, as reflected by significant reductions in BRISQUE and PIQE scores. Moreover, GMFF is capable of performing tasks that these restoration networks cannot, including the removal of edge artifacts, the correction of localized blur, and the enhancement of fine details.

\section{Discussion}
Despite the effectiveness of the proposed framework, it still exhibits four main limitations. First, due to the multi-step sampling inherent in diffusion models, the generative stage suffers from relatively slow processing, which substantially increases the overall runtime. Although the deterministic fusion stage produces an initial fused image in near real-time, the extended inference time of the generative restoration stage diminishes this advantage. Second, constrained by hardware limitations, we employed \textit{Stable Diffusion 2.1-base} as the diffusion prior. As an early text-to-image model, it is prone to errors in rendering textual content, fine facial details, and hand structures. Third, the quality of the fused image is highly dependent on the accuracy of the registration process. Even slight misalignments among the source images can introduce artifacts and significantly degrade the overall fusion quality. Fourth, within the current GMFF framework, the generative restoration stage is essentially treated as a blind image restoration task. However, prior works in the depth-from-focus domain \cite{yang2023aberration,wang2021bridging,won2022learning} have shown that scene depth can be inferred from the original focal stack, suggesting that incorporating stronger priors could provide geometric constraints to further enhance the scene consistency of the generated images.

We envision several promising directions to further advance GMFF: (1) exploring GAN-based generative restoration models as an alternative, as prior studies in HYPIR \cite{lin2025harnessing} indicate that GAN-based models can potentially replace diffusion models while significantly improving inference speed; (2) leveraging more advanced, large-scale pre-trained generative priors, such as SDXL (2.6B parameters) \cite{podell2023sdxl}, SD3 (8B parameters) \cite{esser2024scaling}, and Flux (12B parameters); (3) integrating Depth from Focus as an auxiliary task, allowing the inferred depth and fused images to jointly act as conditional inputs; (4) parameterizing the registration process and embedding it into the network to improve robustness, despite the availability of existing toolkits such as \href{https://github.com/Xinzhe99/OpenFocus}{\textcolor{magenta}{OpenFocus}}, which provides multiple registration algorithms. We believe these directions offer valuable opportunities to enhance both efficiency and quality, providing new insights and a novel paradigm for multi-focus image fusion.

\section{Conclusion}
In this work, we introduced a generative multi-focus fusion framework, GMFF, which operates in two distinct stages: (1) deterministic fusion: We present the latest generation of the StackMFF series, StackMFF V4, which achieves state-of-the-art performance among deterministic multi-focus fusion models. It efficiently fuses the available focal plane information to generate an initial fused image, attaining superior fusion quality with minimal computational overhead; (2) generative restoration: We propose IFControlNet, which leverages the fused output of StackMFF V4 as a conditional input and combines it with a diffusion prior to reconstruct the content of missing focal planes while mitigating common edge artifacts. These two stages are decoupled and can be optimized iteratively and independently. 

Overall, GMFF demonstrates the potential of generative modeling to redefine the paradigm of multi-focus image fusion, bridging deterministic fusion and diffusion-based restoration in a unified and extensible framework.

\bibliographystyle{IEEEtran}
\bibliography{bibliography}

\end{document}